\documentclass[runningheads]{llncs}
\usepackage{graphicx}
\usepackage{amsmath,amssymb} % define this before the line numbering.
\usepackage{color}
\usepackage{wrapfig}
\usepackage{stackengine}
\usepackage{xcolor}
\usepackage{rotating}
\usepackage[export]{adjustbox}
\usepackage{xspace}
\makeatletter
\usepackage{caption}
\newcommand{\mycaption}[2]{\caption{\textbf{#1.}\xspace#2}}

\usepackage{algorithm}
\usepackage[]{algpseudocode}

\usepackage{booktabs}
\usepackage{array, makecell, tabularx}
\usepackage{multirow}

\usepackage[pagebackref=false,breaklinks=true,letterpaper=true,colorlinks,urlcolor=blue,citecolor=blue,linkcolor=blue,bookmarks=false]{hyperref}

\def\etal{\textit{et.~al.}\xspace}

\begin{document}
% \linenumbers
\pagestyle{headings}
\mainmatter

% \setlength{\belowcaptionskip}{-15pt}

 % Replace with your title

\title{Superpixel Sampling Networks}
% Replace with your title

\titlerunning{Superpixel Sampling Networks}
% Replace with a meaningful short version of your title

\authorrunning{Jampani et al.}
% Replace with shorter version of the author list. If there are more authors than fits a line, please use A. Author et al.

\author{Varun Jampani$^1$, Deqing Sun$^1$, Ming-Yu Liu$^1$, \\ Ming-Hsuan Yang$^{1,2}$, Jan Kautz$^1$}

%Please write out author names in full in the paper, i.e. full given and family names. 
%If any authors have names that can be parsed into FirstName LastName in multiple ways, please include the correct parsing, in a comment to the volume editors:
%\index{Lastnames, Firstnames}
%(Do not uncomment it, because you may introduce extra index items if you do that, we will use scripts for introducing index entries...)

\institute{$^1$NVIDIA \hspace{1cm} $^2$UC Merced \\
	\email{ \{vjampani,deqings,mingyul,jkautz\}@nvidia.com, mhyang@ucmerced.edu}
}

\maketitle

\begin{abstract}
Superpixels provide an efficient low/mid-level representation of image data, which greatly reduces the number of image primitives for subsequent vision tasks. Existing superpixel algorithms 
are not differentiable, making them difficult to integrate into otherwise end-to-end trainable deep neural networks. We develop a new differentiable model for superpixel sampling that
leverages deep networks for learning superpixel segmentation. The resulting 
{\em Superpixel Sampling Network} (SSN) is end-to-end trainable, which allows learning task-specific superpixels with flexible loss functions and has fast runtime. Extensive experimental analysis indicates that SSNs not only
outperform existing superpixel algorithms on traditional segmentation benchmarks, but can also learn superpixels for other tasks. In addition, SSNs can be easily integrated into downstream deep networks resulting in performance improvements.

\end{abstract}

\section{Introduction}

% Superpixels in computer vision
Superpixels are an over-segmentation of an image that is formed by grouping image pixels~\cite{ren2003learning} based on 
% color and other 
low-level image properties. They provide a perceptually meaningful tessellation of image content, thereby reducing the number of image primitives for subsequent image processing. Owing to their
representational and computational efficiency, superpixels have become an established low/mid-level image representation and are widely-used in computer vision algorithms such as object detection~\cite{shu2013improving,yan2015object}, 
semantic segmentation~\cite{gould2008multi,sharma2014recursive,gadde16bilateralinception},
saliency estimation~\cite{he2015supercnn,perazzi2012saliency,yang2013saliency,zhu2014saliency},
optical flow estimation~\cite{hu2016highly,lu2013patch,sun2014local,yamaguchi2013robust}, 
depth estimation~\cite{van2013depth}, tracking~\cite{yang2014robust}
to name a few. 
Superpixels are especially widely-used in traditional energy minimization
frameworks, where a low number of image primitives greatly reduce the optimization 
complexity. 
% \DS{the two sentences above are overlapping.}

% Why superpixels are not so popular with deep learning?
The recent years have witnessed a dramatic increase in the adoption of deep learning for
a wide range of computer vision problems. With the exception of a few methods (e.g.,~\cite{gadde16bilateralinception,he2015supercnn,sharma2014recursive}),
superpixels are scarcely used in conjunction with modern deep networks. There are two main reasons for this.
First, the standard convolution operation, which forms the basis of most deep architectures, is usually defined over regular grid lattices and becomes inefficient when operating over irregular superpixel lattices.
Second, existing superpixel algorithms are non-differentiable and thus using superpixels in deep networks introduces non-differentiable modules in otherwise end-to-end trainable network architectures.

\begin{figure}[t]
\begin{center}
\centerline{\includegraphics[width=\columnwidth]{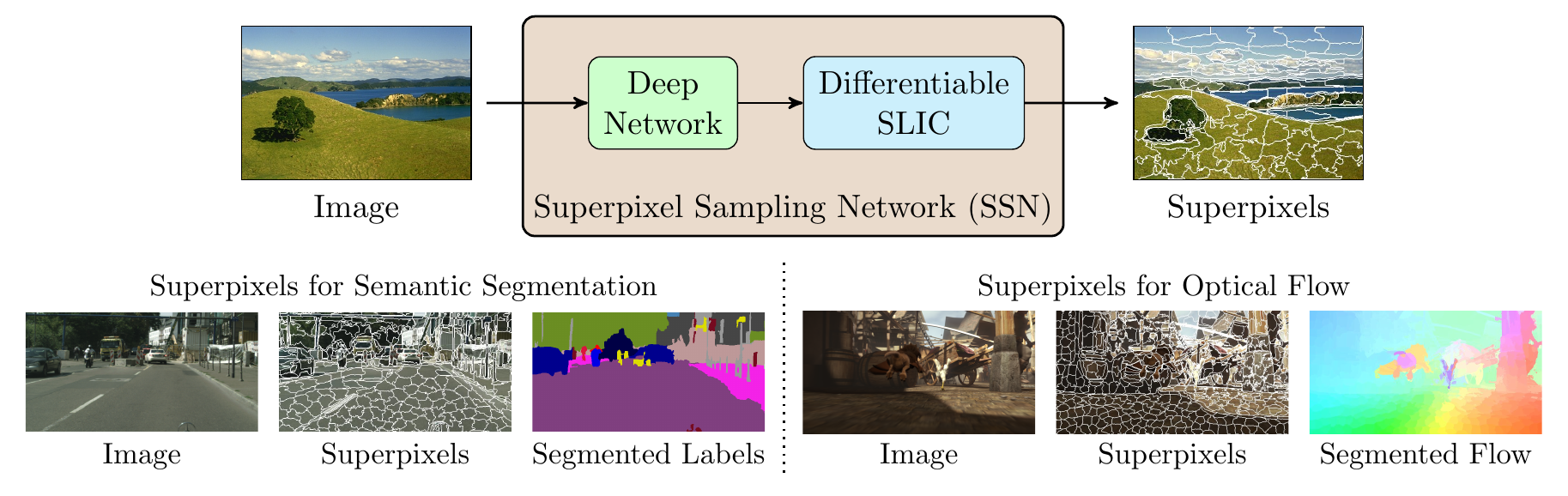}}
  \mycaption{Overview of Superpixel Sampling Networks}{A given image is first passed
  onto a deep network that extracts features at each pixel, which are then used
  by differentiable SLIC to generate the superpixels. Shown here are a couple of example SSN generated task-specific superpixels
  for semantic segmentation and optical flow.}\label{fig:overview}
\end{center}
\end{figure}

% Aim of this work and briefly about the technique
In this work, we alleviate the second issue by proposing a new deep differentiable algorithm for superpixel segmentation. We start by revisiting the widely-used  {\em Simple Linear Iterative Clustering} (SLIC) superpixel algorithm~\cite{achanta2012slic} 
and turn it into a differentiable algorithm by relaxing the nearest neighbor constraints present in SLIC. This new differentiable algorithm allows for end-to-end training and enables us to leverage powerful deep networks for learning superpixels instead of using traditional hand-crafted features. This combination of a deep network with differentiable SLIC forms our end-to-end trainable superpixel algorithm which we call {\em Superpixel Sampling Network} (SSN). 
Fig.~\ref{fig:overview} shows an overview of the proposed SSN. 
A given input image is first passed through a deep network producing features at each pixel. These deep features are then passed onto the differentiable SLIC, which performs iterative clustering, resulting in the desired superpixels. The entire network is end-to-end trainable. The differentiable nature of SSN allows the use of flexible loss functions for learning task-specific superpixels. Fig.~\ref{fig:overview} shows some sample SSN generated superpixels.

% Briefly about the experimental results
Experimental results on 3 different segmentation benchmark datasets including  
BSDS500~\cite{amfm2011bsds},
Cityscapes~\cite{cordts2016cityscapes} and  
PascalVOC~\cite{everingham2015pascal} indicate that the proposed
superpixel sampling network (SSN) performs favourably against existing prominent superpixel algorithms, 
while also being faster. We also demonstrate that by simply integrating our SSN framework into an existing semantic segmentation network~\cite{gadde16bilateralinception} that uses superpixels, performance improvements are achieved. In addition, we demonstrate the flexibility of SSN in learning superpixels for other vision tasks. Specifically, in a proof-of-concept experiment on the Sintel optical flow dataset~\cite{butler2012naturalistic}, we demonstrate how we can learn superpixels that better align with optical flow boundaries rather than standard object boundaries.
% Contributions
The proposed SSN has the following favorable properties in comparison to existing superpixel algorithms:
\begin{itemize}
    \item \textbf{End-to-end trainable:} SSNs are end-to-end trainable and can be easily integrated into other deep network architectures.
    To the best of our knowledge, this is the first end-to-end trainable superpixel algorithm.
    \item \textbf{Flexible and task-specific:} SSN allows for learning with flexible loss functions resulting in the learning of task-specific superpixels.
    \item \textbf{State-of-the-art performance:} Experiments on a wide range of benchmark datasets show that SSN outperforms existing superpixel algorithms.
    \item \textbf{Favorable runtime:} SSN also performs favorably against prominent superpixel algorithms in terms of runtime, making it amenable to learn on large datasets and also effective for practical applications.
\end{itemize}

\section{Related Work}

\noindent \textbf{Superpixel algorithms.}
Traditional superpixel algorithms can be broadly classified into graph-based and clustering-based 
approaches. Graph-based approaches formulate the superpixel segmentation as a graph-partitioning
problem where graph nodes are represented by pixels and the edges denote the strength
of connectivity between adjacent pixels. Usually, the graph partitioning is performed by solving
a discrete optimization problem. 
Some widely-used algorithms in this category include
the normalized-cuts~\cite{ren2003learning}, Felzenszwalb and Huttenlocher 
(FH)~\cite{felzenszwalb2004efficient}, 
and the entropy rate superpixels (ERS)~\cite{liu2011entropy}.
As discrete optimization involves discrete variables, the optimization objectives are
usually non-differentiable making it difficult to leverage deep networks in graph-based
approaches.

% Graph-based approaches model an image using a graph where pixels are represented as graph nodes. 
% The superpixel segmentation task is to select a set of graph edges that form connected components for graph partitioning. Usually, the edge selection is achieved by solving a discrete optimization problem. The widely-used algorithms in this category include the normalized cut superpixel~\cite{ren2003learning}, Felzenszwalb and Huttenlocher (FH)~\cite{felzenszwalb2004efficient}, and the entropy rate superpixel (ERS)~\cite{liu2011entropy}. As the edge selection leads a discrete optimization problem, involving non-differentiable discrete variables, it is difficult to leverage deep networks in these approaches.

Clustering-based approaches, on the other hand, leverage traditional clustering techniques such as
$k$-means for superpixel segmentation. 
Widely-used algorithms in this category include
SLIC~\cite{achanta2012slic}, LSC~\cite{li2015lsc}, and Manifold-SLIC~\cite{liu2016manifold}.
These methods mainly do $k$-means clustering but differ in their feature representation.
While the SLIC~\cite{achanta2012slic} represents each pixel as a $5$-dimensional positional
and \emph{Lab} color features ($XYLab$ features), LSC~\cite{li2015lsc} method projects 
these $5$-dimensional features on to a $10$-dimensional space and performs clustering 
in the projected space.
Manifold-SLIC~\cite{liu2016manifold}, on the other hand, uses a $2$-dimensional manifold
feature space for superpixel clustering.
While these clustering algorithms require iterative updates, a non-iterative clustering scheme for superpixel segmentation is proposed in the SNIC method~\cite{achanta2017snic}.
The proposed approach is also a clustering-based approach. However, unlike existing techniques,
we leverage deep networks to learn features for superpixel clustering via an end-to-end training framework.

% Several widely-used clustering-based superpixel segmentation algorithm include SLIC~\cite{achanta2012slic}, LSC~\cite{li2015lsc}, and Manifold-SLIC~\cite{liu2016manifold}. 
% These methods are all based on the $k$-mean clustering but differ in their feature representation. While the SLIC~\cite{achanta2012slic} represents each pixel as a $5$-dimensional vector consists of its XY coordinates and CIELab color, the LSC~\cite{li2015lsc} method projects the $5$-dimensional features to a 10 dimensional space and performs clustering in the projected space. The Manifold-SLIC~\cite{liu2016manifold} method uses a 2-dimensional feature representation. While these clustering algorithms require iterative updates, a non-iterative clustering scheme for superpixel segmentation is proposed in the SNIC work~\cite{achanta2017snic}. 
% The proposed method belongs to this category. But instead of using hand-crafted features, we learn deep representation via an end-to-end training framework.

As detailed in a recent survey paper~\cite{stutz2016benchmark}, other techniques are used for superpixel segmentation, including watershed transform~\cite{machairas2015waterpixels}, geometric flows~\cite{levinshtein2009turbopixels}, graph-cuts~\cite{veksler2010superpixels}, mean-shift~\cite{comaniciu2002mean}, and hill-climbing~\cite{van2015seeds}. However, these methods all
rely on hand-crafted features and it is non-trivial to incorporate deep networks into these techniques. A very recent technique of SEAL~\cite{Tu-CVPR-2018} proposed a way to learn deep features for superpixel segmentation by bypassing the gradients through non-differentiable superpixel algorithms. Unlike our SSN framework,
SEAL is not end-to-end differentiable.

%Interested readers could refer to a recent survey of superpixels~\cite{stutz2016benchmark}.

\noindent \textbf{Deep clustering.}
Inspired by the success of deep learning for supervised tasks, several methods investigate the use of deep networks for unsupervised data clustering. Recently, Greff \etal~\cite{greff2017neural} propose the neural expectation maximization framework where they model the posterior distribution of cluster labels using deep networks and unroll the iterative steps in the EM procedure for end-to-end training.
In another work~\cite{greff2016tagger}, the Ladder network~\cite{rasmus2015semi} is used to model a hierarchical latent variable model for clustering. Hershey \etal~\cite{hershey2016deep} propose a deep learning-based clustering framework for separating and segmenting audio signals. Xie \etal \cite{xie2016unsupervised} propose a deep embedded clustering framework, for simultaneously learning feature representations and cluster assignments. 
%MH: people et al. manes a copuple people. I do not treat it as a paper...
%and so use "give" not "gives"
In a recent survey paper, Aljalbout~\etal \cite{aljalbout2018clustering} give a taxonomy of deep learning based clustering methods. 
In this paper, we also propose a deep learning-based clustering algorithm. Different 
from the prior work, our algorithm is tailored for the superpixel segmentation task where we use image-specific constraints. Moreover, our framework can easily incorporate other vision objective functions for learning task-specific superpixel representations.

%\section{Preliminaries of the SLIC Superpixel Algorithm}
\section{Preliminaries}
\label{sec:slic_review}

% Why SLIC? And, its importance in superpixel literature.
At the core of SSN is a differentiable clustering technique that is inspired by the
SLIC~\cite{achanta2012slic} superpixel algorithm. Here, we briefly review the SLIC 
before describing our SSN technique in the next section.
SLIC is one of the simplest and
also one of the most widely-used superpixel algorithms. It is easy to implement,
has fast runtime and also produces compact and uniform superpixels.
% that adhere to the object boundaries.

% Brief explanation of SLIC superpixel algorithm
Although there are several different variants~\cite{li2015lsc,liu2016manifold} of SLIC algorithm, in the original
form, SLIC is a k-means clustering performed on image pixels in a five dimensional
position and color space (usually scaled $XYLab$ space).
Formally, given an image $I \in \mathbb{R}^{n \times 5}$, with $5$-dimensional
$XYLab$ features at $n$ pixels, the task of superpixel computation is to assign each
pixel to one of the $m$ superpixels i.e., to compute the pixel-superpixel
association map $H \in \{0,1,\cdots,m-1\}^{n \times 1}$.
The SLIC algorithm operates as follows. First, we sample 
initial cluster (superpixel) centers $S^0 \in \mathbb{R}^{m \times 5}$ in the 
$5$-dimensional space. This sampling is usually done
uniformly across the pixel grid with some local perturbations based on image gradients.
Given these initial superpixel centers $S^0$, the SLIC algorithm proceeds in an iterative manner
with the following two steps in each iteration $t$:
\begin{enumerate}
    \item \textit{Pixel-Superpixel association}: Associate each pixel to the nearest superpixel
    center in the five-dimensional space, i.e., compute the new superpixel assignment at each
    pixel $p$, 
    \begin{equation}
        H_p^t = \underset{i \in \{0,...,m-1\}}{\arg \min}D(I_p, S^{t-1}_i),
        \label{eqn:slic_1}
    \end{equation}
    where $D$ denotes the distance computation $D(\textbf{a},\textbf{b}) = ||\textbf{a}-\textbf{b}||^2$.
    \item \textit{Superpixel center update}: Average pixel features ($XYLab$) inside each
    superpixel cluster to obtain new superpixel cluster centers $S^t$. For each superpixel~$i$,
    we compute the centroid of that cluster,
    \begin{equation}
        S^t_i = \frac{1}{Z_i^t}\sum_{p | H_p^t = i} I_p,
        \label{eqn:slic_2}
    \end{equation}
    where $Z_i^t$ denotes the number of pixels in the superpixel cluster $i$.
\end{enumerate}
These two steps form the
core of the SLIC algorithm and are repeated until either convergence or for a fixed 
number of iterations. Since computing the distance $D$ in Eq.~\ref{eqn:slic_1} 
between all the pixels and superpixels is time-consuming, this computation is usually constrained
to a fixed neighborhood around each superpixel center.
At the end, depending on the application, there is an optional step
of enforcing spatial connectivity across pixels in each superpixel cluster. 
More details regarding the SLIC algorithm can be found in Achanta \etal~\cite{achanta2012slic}.
In the next
section, we elucidate how we modify the SLIC algorithm to develop SSN.

\section{Superpixel Sampling Networks}

As illustrated in Fig.~\ref{fig:overview}, SSN is composed of two parts: A deep network
that generates pixel features, which are then passed on to differentiable SLIC. 
Here, we first describe the differentiable SLIC followed by the 
SSN architecture.

\subsection{Differentiable SLIC}
\label{sec:dslic}

% Why is SLIC not differentiable?
\textit{Why is SLIC not differentiable?} A closer look at all the computations
in SLIC shows that the non-differentiability arises because of the
computation of pixel-superpixel associations, which involves a non-differentiable
nearest neighbor operation. This nearest neighbor computation also forms the
core of the SLIC superpixel clustering and thus we cannot avoid this operation.

A key to our approach is to convert the nearest-neighbor operation into a
differentiable one. Instead of computing hard pixel-superpixel 
associations $H \in \{0,1,\cdots,m-1\}^{n \times 1}$ (in Eq.~\ref{eqn:slic_1}),
we propose to compute soft-associations
$Q \in \mathbb{R}^{n \times m}$ between pixels and superpixels. 
Specifically, for a pixel $p$ and superpixel $i$ at iteration $t$,
we replace the nearest-neighbor computation (Eq.~\ref{eqn:slic_1}) in
SLIC with the following pixel-superpixel association.

\begin{equation}
    Q_{pi}^t = e^{-D(I_p,S_i^{t-1})} = e^{-||I_p - S_i^{t-1}||^2}
    \label{eqn:ssn_1}
\end{equation}
% \DS{should there be a scaling factor, i.e. exp(-alpha *)?}

\begin{figure}[t]
\begin{center}
\centerline{\includegraphics[width=\columnwidth]{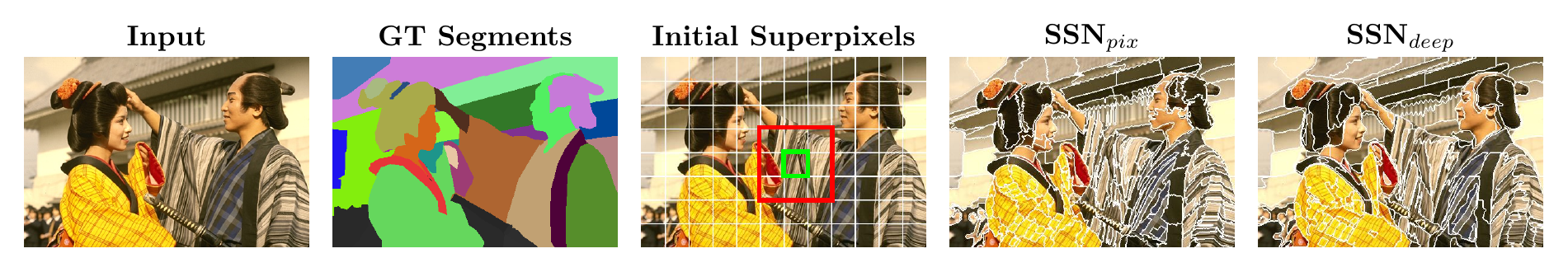}}
  \mycaption{From initial grid to learned superpixels}{An example visual result from BSDS500 dataset
  showing the initial superpixel grid and the superpixels obtained with SSN$_{pix}$ and SSN$_{deep}$.
  To compute the pixel-superpixel associations for every pixel in the green box, only the surrounding
  superpixels in the red box are considered.}
  \label{fig:bsds_illustration}
\end{center}
\end{figure}

Correspondingly, the computation of new superpixels cluster centers 
(Eq.~\ref{eqn:slic_2}) is modified as the weighted sum of pixel features,

\begin{equation}
     S^t_i = \frac{1}{Z_i^t}\sum_{p=1}^n Q_{pi}^t I_p,
    \label{eqn:ssn_2}
\end{equation}
where $Z_i^t = \sum_p Q_{pi}^t$ is the normalization constant. For convenience, we refer
to the column normalized $Q^t$ as $\hat{Q^t}$ and thus we can write the above superpixel
center update as $S^t = \hat{Q^t}^{\top} I$. The size of $Q$ is $n \times m$ and even
for a small number of superpixels $m$, it is prohibitively expensive to compute $Q_{pi}$
between all the pixels and superpixels. Therefore, we constrain the distance computations
from each pixel to only 9 surrounding superpixels as illustrated using the red and green
boxes in Fig.~\ref{fig:bsds_illustration}. For each pixel in the green box, only the surrounding
superpixels in the red box are considered for computing the association.
This brings down the size of $Q$ from $n \times m$ to $n \times 9$,
making it efficient in terms of both computation
and memory. This approximation in the $Q$ computation is similar in spirit
to the approximate nearest-neighbor search in SLIC.
% \DS{We may experiment with the nubmer of surrounding superpixels}

\begin{algorithm}[t]
\caption{Superpixel Sampling Network (SSN)}
\label{alg:ssn}
\begin{algorithmic}[1]
\Statex \textbf{Input:} Image $\underset{n \times 5}{I}$. \Comment{$XYLab$ features} 
\Statex \textbf{Output:} Pixel-Superpixel association $\underset{n \times m}{Q}$.
\State Pixel features using a CNN, $\underset{n \times k}{F} = \mathcal{F}(I)$.
\State Initial superpixel centers with average features in regular grid cells, $\underset{m \times k}{S^0} = \mathcal{J}(F)$.
\For{each iteration $t$ in $1$ to $v$}
    \State Compute association between each pixel $p$ and the surrounding superpixel $i$, $Q_{pi}^t = e^{-||F_p - S_i^{t-1}||^2}$.
    \State Compute new superpixel centers, $S^t_i = \frac{1}{Z_i^t}\sum_{p=1}^n Q_{pi}^t F_p$; $Z_i^t = \sum_p Q_{pi}^t$.
    
\EndFor
\State (\textit{Optional}) Compute hard-associations $\underset{n \times 1}{H^v}; 
H_p^v = \underset{i \in \{0,...,m-1\}}{\arg \max}Q_{pi}^v$.
\State (\textit{Optional}) Enforce spatial connectivity.
\end{algorithmic}
\end{algorithm}

Now, both the computations in each SLIC iteration are completely differentiable and we
refer to this modified algorithm as {\em differentiable SLIC}. 
Empirically, we observe that replacing the hard pixel-superpixel associations in SLIC
with the soft ones in differentiable SLIC does not result in any performance degradations. 
Since this new superpixel algorithm
is differentiable, it can be easily integrated into any deep network architecture.
Instead of using manually designed pixel features $I_p$, we can leverage deep feature
extractors and train the whole network end-to-end. In other words, we replace the image
features $I_p$ in the above computations (Eq.~\ref{eqn:ssn_1} and~\ref{eqn:ssn_2})
with $k$ dimensional pixel features $F_p \in \mathbb{R}^{n \times k}$ 
computed using a deep network. We refer to this coupling of deep networks with the
differentiable SLIC as {\em Superpixel Sampling Network} (SSN). 

Algorithm~\ref{alg:ssn} outlines all the computation steps in SSN. The algorithm starts with
deep image feature extraction using a CNN (line 1). We initialize the superpixel
cluster centers (line 2)
with the average pixels features in an initial regular superpixel grid (Fig.~\ref{fig:bsds_illustration}). 
Then, for $v$ iterations, we iteratively update pixel-superpixel associations and superpixel
centers, using the above-mentioned computations (lines 3-6).
Although one could directly use soft pixel-superpixel associations $Q$ for several downstream
tasks, there is an optional step of converting soft associations to hard ones (line 7), depending on the application needs. 
% This also enables comparisons with other superpixel techniques.
In addition, like in the original SLIC algorithm, we can optionally enforce spatial connectivity
across pixels inside each superpixel cluster. This is accomplished by merging the superpixels,
smaller than certain threshold, with the surrounding ones and then assigning a unique cluster ID
for each spatially-connected component. Note that these two optional steps (lines 7, 8) are not
differentiable.
% \JK{say that this part (and previous) is not differentiable?}

\begin{figure}[t]
\begin{center}
\centerline{\includegraphics[width=\columnwidth]{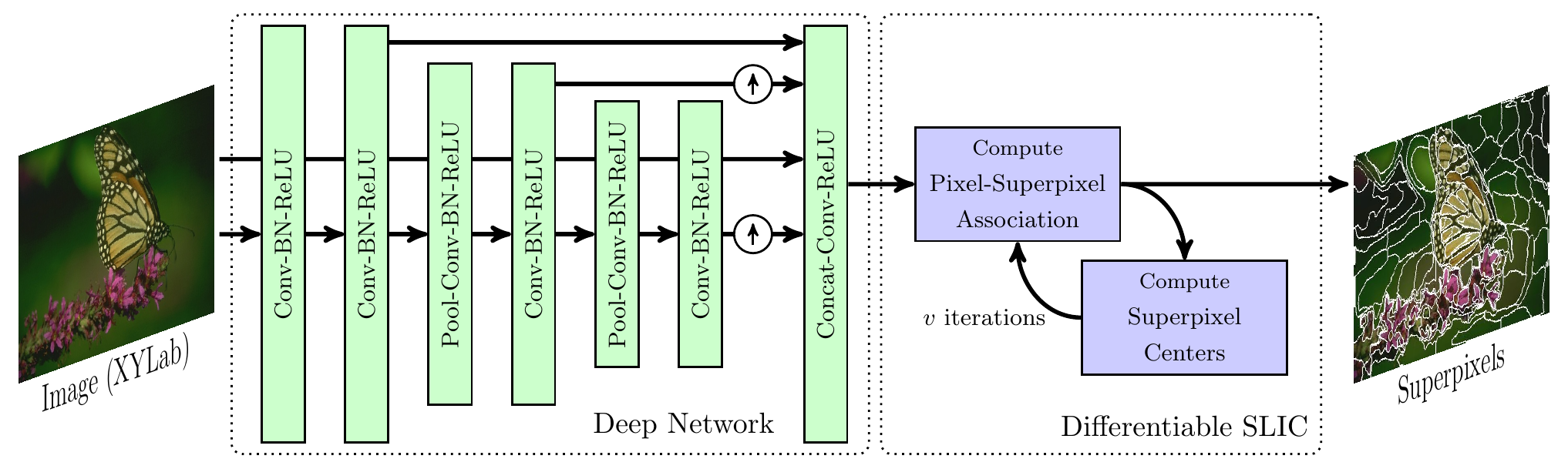}}
  \mycaption{Computation flow of SSN}{Our network is composed of a series of convolution layers interleaved with Batch Norm (BN) and ReLU nonlinearities. $\uparrow$ denotes bilinear upsampling to the original image resolution. The features from CNNs are then passed onto iterative updates in the differentiable SLIC to generate superpixels.}\label{fig:netarch}
\end{center}
\end{figure}

\noindent \textbf{Mapping between pixel and superpixel representations.} For some downstream applications
that use superpixels, pixel representations are mapped onto superpixel representations
and vice versa. With the traditional superpixel algorithms, which provide hard clusters, this
mapping from pixel to superpixel representations is done via averaging inside each cluster 
(Eq.~\ref{eqn:slic_2}). The inverse mapping from superpixel to pixel representations is done by
assigning the same superpixel feature to all the pixels belonging to that superpixel.
We can use the same pixel-superpixel mappings with SSN superpixels as well, using the
hard clusters (line 7 in Algorithm~\ref{alg:ssn}) obtained from SSN.
However, since this computation of hard-associations is not differentiable, it may not be desirable
to use hard clusters when integrating into an end-to-end trainable system. 
It is worth noting that the soft pixel-superpixel associations generated by SSN can also be easily used for mapping between pixel and superpixel representations. Eq.~\ref{eqn:ssn_2} already
describes the mapping from a pixel to superpixel representation
which is a simple matrix multiplication with the transpose of column-normalized $Q$ matrix: 
$S = \hat{Q}^{\top}F$, where $F$ and $S$ denote pixel and superpixel representations respectively.
The inverse mapping from superpixel to pixel representation is done by multiplying
the row-normalized $Q$, denoted as $\tilde{Q}$, with the superpixel representations,
$F = \tilde{Q} S$. Thus the pixel-superpixel feature mappings are
given as simple matrix multiplications with the association matrix and are differentiable.
Later, we will make use of these mappings in designing the loss functions to train SSN.

\subsection{Network Architecture}

Fig.~\ref{fig:netarch} shows the SSN network architecture. 
The CNN for feature extraction
is composed of a series of convolution layers interleaved with 
batch normalization~\cite{ioffe2015batch} (BN)
and ReLU activations. We use max-pooling, which downsamples the input 
by a factor of 2, after the 2$^{nd}$ and 4$^{th}$ convolution layers to increase the receptive
field. We bilinearly upsample the 4$^{th}$ and 6$^{th}$ convolution layer outputs 
and then concatenate with the 2$^{nd}$ convolution layer output to pass
onto the final convolution layer.
We use $3 \times 3$ convolution filters with
the number of output channels set to 64 in each layer, 
except the last CNN layer which outputs $k-5$ channels.
We concatenate this $k-5$ channel output with the $XYLab$ of the given image resulting
in $k$-dimensional pixel features.
We choose this CNN architecture for its simplicity and efficiency. Other network 
architectures are conceivable. 
The resulting $k$ dimensional features are passed onto the two modules of differentiable SLIC
that iteratively updates pixel-superpixel associations and superpixel centers for $v$ iterations.
The entire network is end-to-end trainable.

\subsection{Learning Task-Specific Superpixels}

One of the main advantages of end-to-end trainable SSN is the flexibility in terms of loss 
functions, which we can use to learn task-specific superpixels. 
Like in any CNN, we can couple SSN with
any task-specific loss function resulting in the learning of superpixels that are
optimized for downstream computer vision tasks.
In this work, we focus on optimizing the representational
efficiency of superpixels i.e., learning superpixels that can 
efficiently represent a scene
characteristic such as semantic labels, optical flow, depth etc. 
As an example, if we want to learn superpixels that are going to be used for 
downstream semantic segmentation task, it is desirable to produce superpixels that
adhere to semantic boundaries. To optimize for representational efficiency, 
we find that the combination of a task-specific reconstruction loss and a compactness loss performs well.

\noindent \textbf{Task-specific reconstruction loss.} We denote the pixel properties  
that we want to represent
efficiently with superpixels as $R \in \mathbb{R}^{n \times l}$. 
For instance, $R$ can be semantic
label (as one-hot encoding) or optical flow maps. 
It is important to note that we do not have access to $R$ during the test time, i.e., SSN predicts
superpixels only using image data. 
We only use $R$ during training
so that SSN can learn to predict superpixels suitable to represent $R$.
As mentioned previously in Section~\ref{sec:dslic}, 
we can map the pixel properties onto superpixels using the column-normalized 
association matrix $\hat{Q}$, $\breve{R} = \hat{Q}^{\top}R$, where 
$\breve{R} \in \mathbb{R}^{m \times l}$. The resulting superpixel representation $\breve{R}$ 
is then mapped back onto pixel representation $R^*$ using
row-normalized association matrix $\tilde{Q}$, $R^* = \tilde{Q} S$, 
where $R^* \in \mathbb{R}^{n \times l}$. Then the reconstruction loss is given as

\begin{equation}
    L_{recon} = \mathcal{L}(R, R^*) = \mathcal{L}(R, \tilde{Q} \hat{Q}^{\top}R) % ||R - R^*||_p,
    \label{eqn:recon}
\end{equation}
where $\mathcal{L}(.,.)$ denotes a task-specific loss-function.
% \DS{how about expanding $R^*$ in the reconstruction loss, i.e., $\mathcal{L}(R, \tilde{Q} \hat{Q}^{\top}R)$ ?}
In this work, for segmentation tasks, we used cross-entropy loss for $\mathcal{L}$
and used L1-norm for learning superpixels for optical flow.
% $p$-norm for regression tasks and
% where $||.||_p$ denotes $p$-norm. We observe that either L1 or L2 norm works well in practice.
Here $Q$ denotes the association matrix $Q^v$ after the final iteration of differentiable SLIC.
We omit $v$ for convenience.

\noindent \textbf{Compactness loss.} In addition to the above loss,
we also  use a compactness loss to encourage superpixels to be spatially compact i.e.,
to have lower spatial variance inside each superpixel cluster. Let $I^{xy}$ denote positional
pixel features. We first map these positional features into our superpixel representation,
$S^{xy} = \hat{Q}^{\top}I^{xy}$. Then, we do the inverse mapping onto the pixel representation
using the hard associations $H$, instead of soft associations $Q$, by assigning the same
superpixel positional feature to all the pixels belonging to that superpixel,
$\bar{I}_p^{xy} = S_i^{xy} | H_p = i$. The compactness loss is defined as the following L2 norm:

\begin{equation}
    L_{compact} = ||I^{xy} - \bar{I}^{xy} ||_2.
    \label{eqn:compact}
\end{equation}

% \DS{Is this hard assignment differentiable w.r.t. Q, which H seems to depend on?}
This loss encourages superpixels to have lower spatial variance.
% Note that this compactness loss in not differentiable w.r.t. $H$ due to hard-assignments,
% but is differentiable w.r.t. superpixel features $S_i^{xy}$ and thus still plays a role
% in learning the deep network.
The flexibility of SSN allows using many other loss functions,
which makes for interesting future research.
The overall loss we use in this work is a combination of these two loss functions,
$L = L_{recon} + \lambda L_{compact}$, where we set $\lambda$ to $10^{-5}$ in all our
experiments.
% \DS{The reconstruction losses may be at different scales for different applications; thus $lambda=1e-5$ may not be optimal, right?}

\subsection{Implementation and Experiment Protocols}

We implement the differentiable SLIC as neural network layers using CUDA in the Caffe neural
network framework~\cite{jia2014caffe}. All the experiments are performed using Caffe 
with the Python interface. 
% Scaling of XYLab
We use scaled $XYLab$ 
features as input to the SSN, with position and color feature scales represented
as $\gamma_{pos}$ and $\gamma_{color}$ respectively. The value of $\gamma_{color}$ 
is independent of the number of superpixels and is set to 0.26 with
color values ranging between 0 and 255. The value of $\gamma_{pos}$ depends on the
number of superpixels, $\gamma_{pos} = \eta \max{(m_w/n_w, m_h/n_h)}$, 
where $m_w,n_w$ and $m_h,n_h$ denotes
the number of superpixels and pixels along the image width and height respectively. 
In practice, we observe that $\eta = 2.5$ performs well.

% Training details
For training, we use image patches of size $201 \times 201$ and
$100$ superpixels. In terms of data augmentation, we use left-right flips and for the
small BSDS500 dataset~\cite{amfm2011bsds}, 
we use an additional data augmentation of random
scaling of image patches. For all the experiments, we use
Adam stochastic optimization~\cite{kingma2014adam} with a
batch size of $8$ and a learning rate of $0.0001$. Unless otherwise mentioned,
we trained the models for 500K iterations and choose the final trained models
based on validation accuracy. 
% In cases, where there is no separate validation
% and test sets, we use a small subset of train
% % \JK{? I thought there isn't a separate train and validation set?} 
% and validation images for validating
% the trained models and evaluate on the entire validation set. 
For the ablation studies, we trained models with varying parameters for 200K iterations. 
It is important to note that we use a single trained SSN model
for estimating varying number of superpixels by scaling the input positional
features as described above. 
We use 5 iterations ($v=5$) of differentiable SLIC for training and used 10 iterations
while testing as we observed only marginal performance gains with more iterations.
% We use 20 dimensional deep features ($k=20$) for all the experiments.
Refer to \url{https://varunjampani.github.io/ssn/} for the code and trained models.

\section{Experiments}
\label{sec:exps}

We conduct experiments on 4 different benchmark datasets. We first demonstrate the use of
learned superpixels with experiments on the prominent superpixel benchmark BSDS500~\cite{amfm2011bsds} 
(Section~\ref{sec:bsds}). We then demonstrate the use of task-specific superpixels on the
Cityscapes~\cite{cordts2016cityscapes} and PascalVOC~\cite{everingham2015pascal} datasets for semantic segmentation (Section~\ref{sec:semseg}),
and on MPI-Sintel~\cite{butler2012naturalistic} dataset for optical flow (Section~\ref{sec:flow}).
In addition, we demonstrate the use of SSN superpixels in a downstream semantic segmentation
network that uses superpixels (Section~\ref{sec:semseg}).

\subsection{Learned Superpixels}
\label{sec:bsds}

We perform ablation studies and evaluate against other superpixel techniques on
the BSDS500 benchmark dataset~\cite{amfm2011bsds}. 
BSDS500 consists of 
200 train, 100 validation, and 200 test images. Each image is annotated
with ground-truth (GT) segments from multiple annotators. 
We treat each annotation as
as a separate sample resulting in 1633 training/validation pairs and
1063 testing pairs. 

In order to learn superpixels that adhere to GT segments, 
we use GT segment labels in the reconstruction loss 
(Eq.~\ref{eqn:recon}). Specifically, we represent GT segments in each
image as one-hot encoding vectors and use that as pixel properties $R$ 
in the reconstruction loss.
We use the cross-entropy loss for $\mathcal{L}$ in Eq.~\ref{eqn:recon}. 
Note that, unlike in the semantic segmentation task where the GT labels 
have meaning, GT segments in this dataset
do not carry any semantic meaning.
% and thus the one-hot encoded labels do not
% represent the same semantic segments across different images. 
%
This does not pose any
issue to our learning setup as both the SSN and reconstruction loss are 
agnostic to the meaning of pixel properties $R$. 
%
% SSN takes images as input and produces superpixel clusters. 
%
The reconstruction loss generates a loss value using the given input signal $R$
and its reconstructed version $R^*$ and does not consider whether the meaning
of $R$ is preserved across images.

\noindent \textbf{Evaluation metrics.} Superpixels are useful in a wide range of
vision tasks and several metrics exist for evaluating superpixels.
In this work, we consider Achievable Segmentation Accuracy (ASA) as our primary
metric while also reporting boundary metrics such as Boundary Recall (BR)
and Boundary Precision (BP) metrics. 
% We use two standard superpixel segementation
% metrics of Achievable  Segmentation Accuracy (ASA) and Boundary Recall (BR) 
% to analyze the performance of generated superpixels. 
%
ASA score represents the upper bound on the accuracy achievable by any segmentation
step performed on the superpixels. Boundary precision and recall on the other
hand measures how well the superpixel boundaries align with the GT boundaries.
We explain these metrics in more detail in the supplementary material.
% To compute ASA score,
% we first map the GT segment labels into superpixel labels, by taking the
% highest occuring label inside each superpixel. We then reconstruct pixel segments from
% the superpixel labels by assigning the same superpixel label to all the pixels
% belonging to a superpixel. Then the ASA score is computed as the pixel
% accuracy of the reconstructed segments with respect to the GT segments.
% BR score, on the other hand, measures the superpixel boundary adherence with
% respect to the GT boundaries. BR score represents the count of superpixel
% boundaries that fall within a close range of GT boundaries. 
% We set the boundary tolerance to 2 pixels while computing BR scores.
% %
% A formal definition of these metrics is included in the supplementary
% material. 
%
The higher these scores, the better is the segmentation result.
We report the average ASA and boundary metrics by varying the 
average number of generated superpixels. 
A fair evaluation of boundary precision and recall expects superpixels to be 
spatially connected. 
Thus, for the sake of unbiased comparisons, we follow the optional
post-processing of computing hard clusters and enforcing spatial connectivity
(lines 7--8 in Algorithm~\ref{alg:ssn}) on SSN superpixels.

\noindent \textbf{Ablation studies.} 
% We conduct some ablation studies with varying network
% architecture, dimensionality of the deep features ($k$) and the number of
% iterations $v$ in the differentiable SLIC. 
We refer to our main model illustrated
in Fig.~\ref{fig:netarch}, with 7 convolution layers in deep network, as
SSN$_{deep}$.
As a baseline model, we evalute the superpixels generated 
with differentiable SLIC that takes pixel $XYLab$ features as input. 
This is similar to standard SLIC algorithm, which we
refer to as SSN$_{pix}$ and has no trainable parameters. As an another baseline
model, we replaced the deep network with a single convolution layer that learns
to linearly transform input $XYLab$ features, which we refer to as SSN$_{linear}$.

\begin{figure}[t]
\footnotesize
\hspace{0.75cm}%
\stackunder[0pt]{\includegraphics[width=5.0cm]{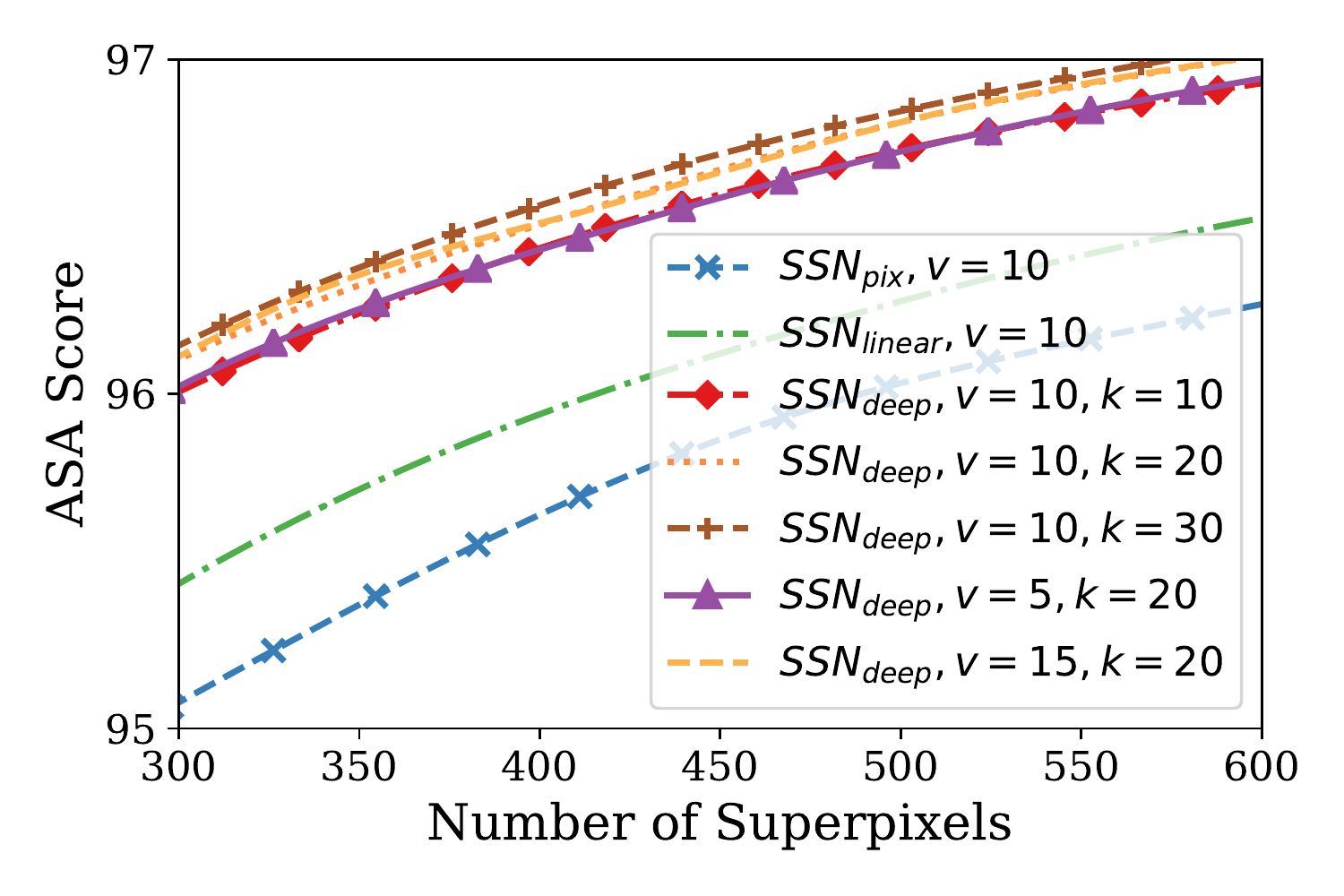}}{}%
\hspace{0.5cm}%
\stackunder[0pt]{\includegraphics[width=5.0cm]{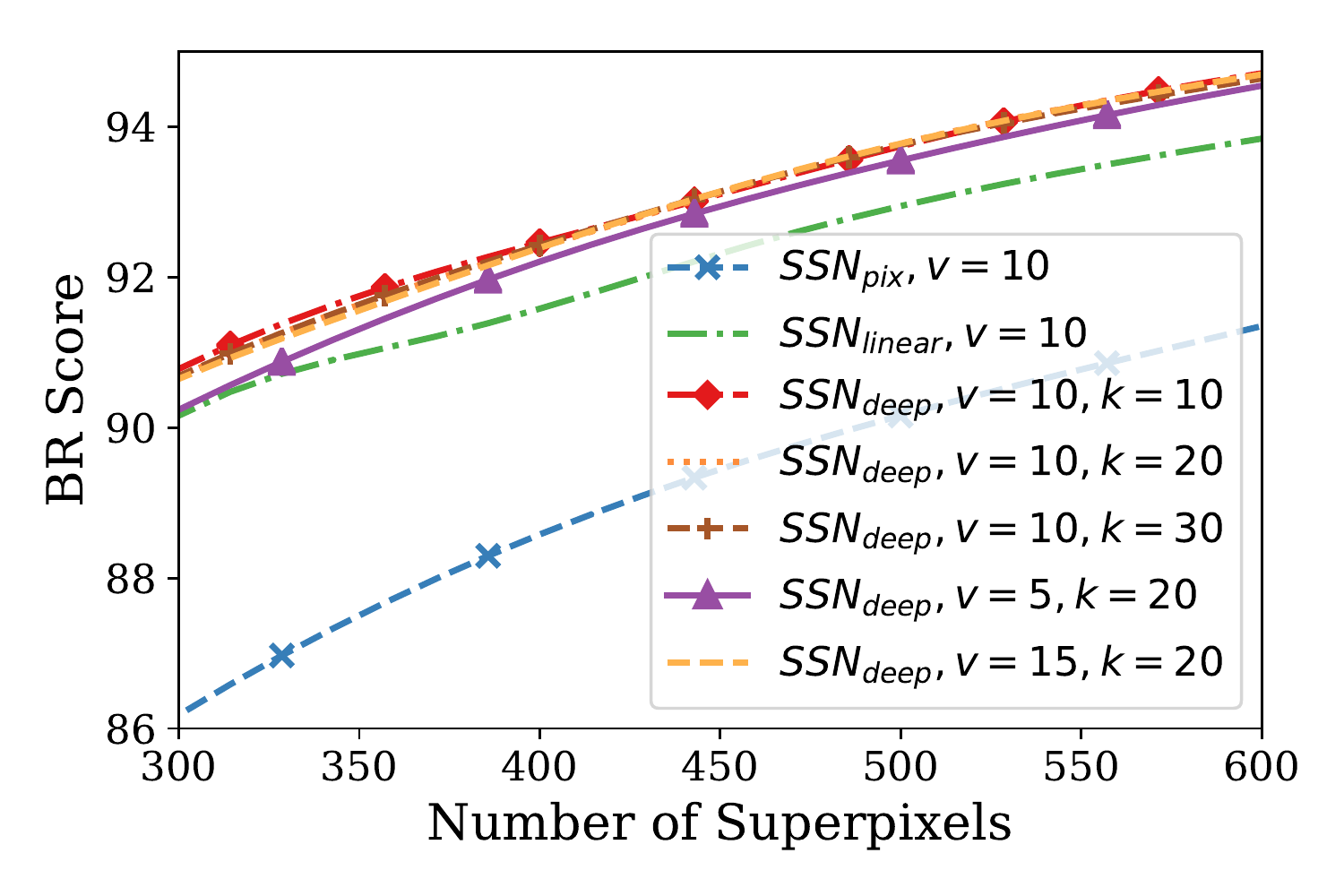}}{}
\mycaption{Ablation studies on BSDS500}{Results on the test set show that 
both the ASA and BR scores considerably 
improve with deep network, and marginally improve with higher number
of feature dimensions $k$ and differentiable SLIC iterations $v$.}
\label{fig:ablation}
\end{figure}

Fig.~\ref{fig:ablation} shows the average ASA and BR scores for these different
models with varying feature dimensionality $k$ and the number of iterations $v$
in differentiable SLIC. The ASA and BR of SSN$_{linear}$ is already reliably
higher than the baseline SSN$_{pix}$ showing the importance of our loss functions
and back-propagating the loss signal through the superpixel algorithm. SSN$_{deep}$
% with the help of the deep network, 
further improves ASA and BR scores by a large
margin. We observe slightly better scores with higher feature dimensionality $k$
and also more iterations $v$. For computational reasons, we choose
$k=20$ and $v=10$ and from here on refer to this model as SSN$_{deep}$.

\begin{figure}[h]
\footnotesize
\hspace{0.75cm}%
\stackunder[0pt]{\includegraphics[width=5.0cm]{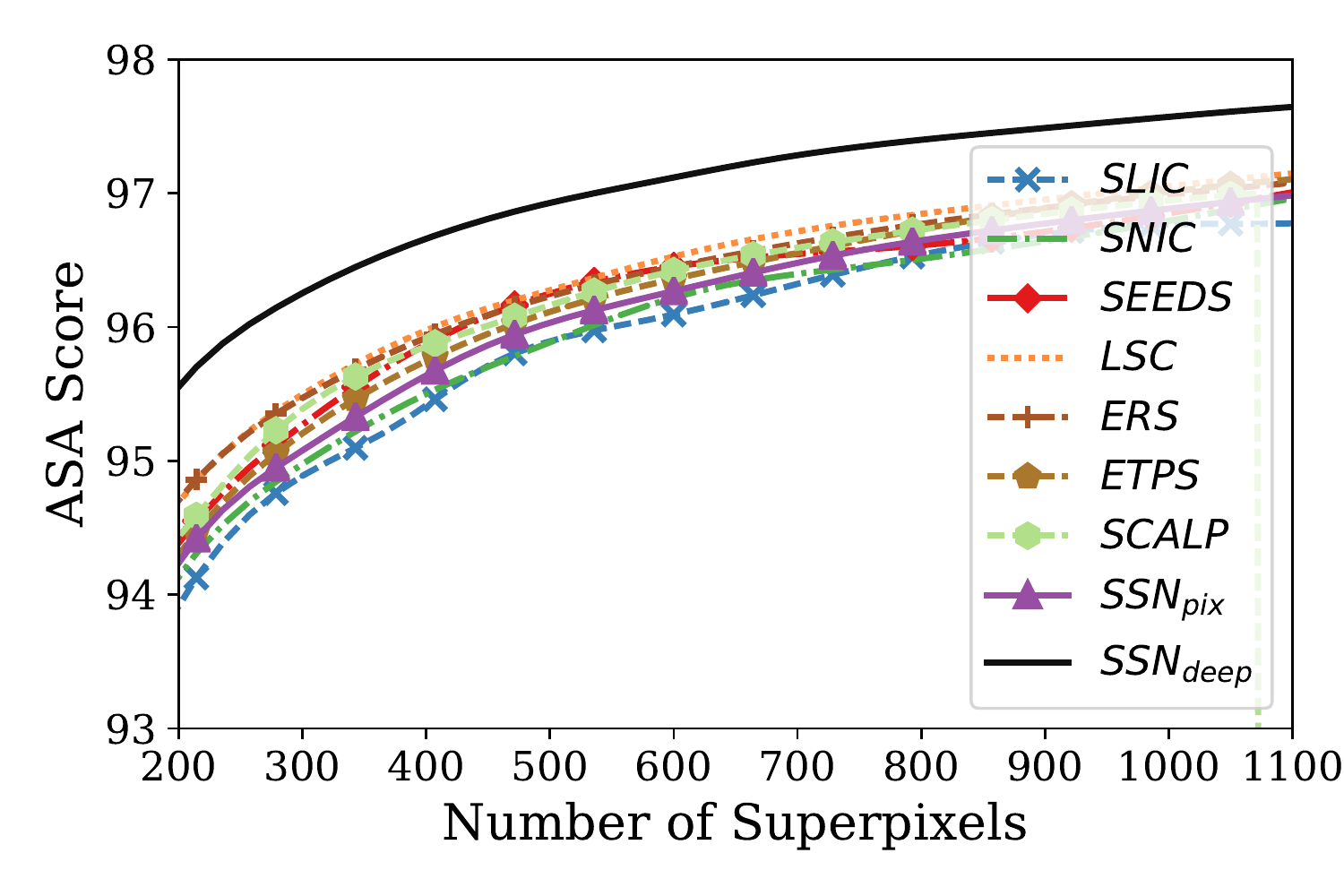}}{}%
\hspace{0.5cm}%
\stackunder[0pt]{\includegraphics[width=5.0cm]{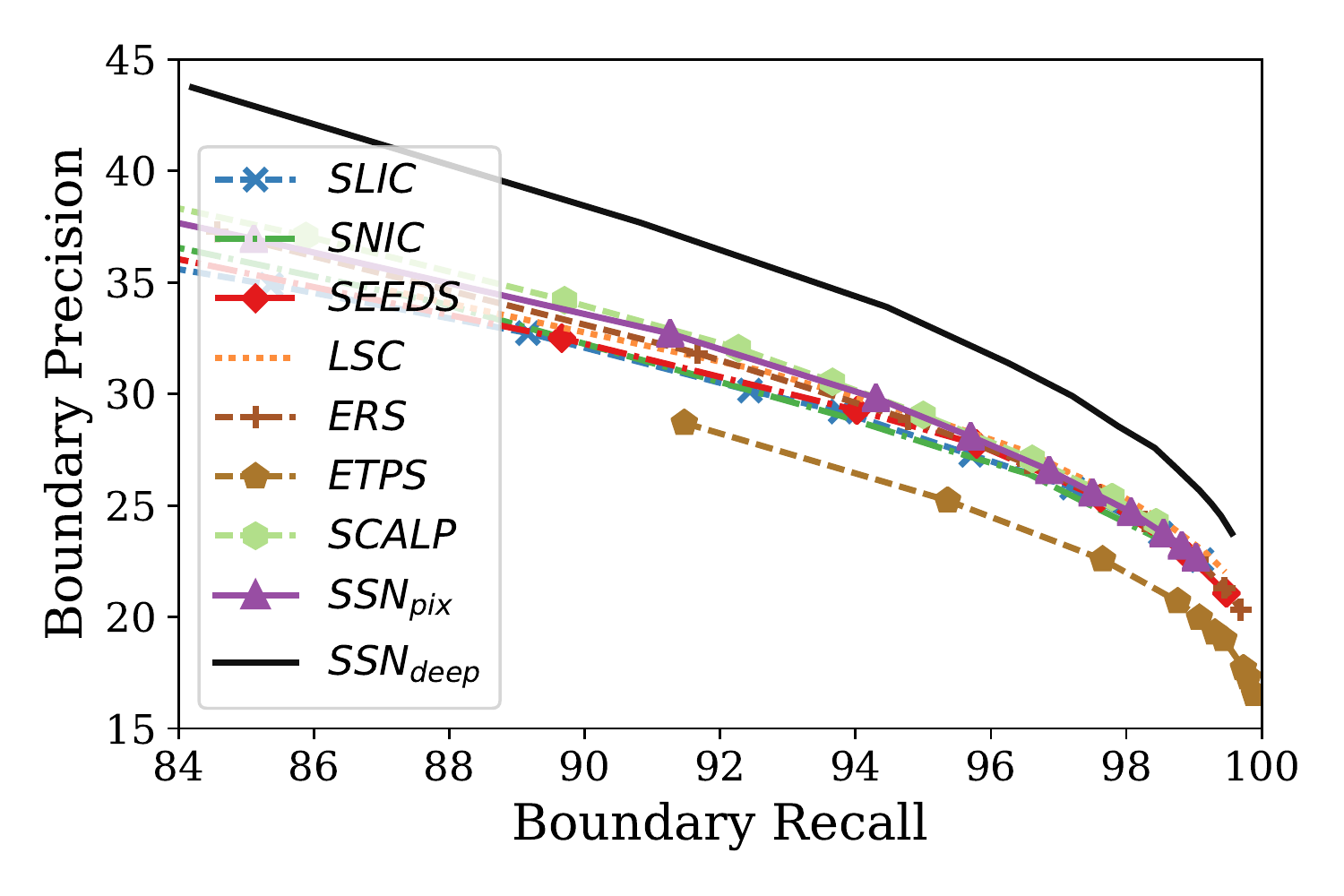}}{}
\mycaption{Results on BSDS500 test}{SSN performs favourably against other techniques in 
terms of both ASA score and boundary precision-recall.}
\label{fig:scores_bsds}
\end{figure}

\noindent \textbf{Comparison with the state-of-the-arts.} Fig.~\ref{fig:scores_bsds}
shows the ASA and precision-recall comparison of SSN with state-of-the-art 
superpixel algorithms. We compare with the following prominent algorithms:
SLIC~\cite{achanta2012slic}, SNIC~\cite{achanta2017snic}, 
SEEDS~\cite{van2015seeds}, LSC~\cite{li2015lsc}, ERS~\cite{liu2011entropy},
ETPS~\cite{yao2015real} and SCALP~\cite{giraud2016scalp}.
Plots indicate that SSN$_{pix}$ performs similarly to
SLIC superpixels, showing that the performance of SLIC does not drop
when relaxing the nearest neighbor constraints. Comparison with other
techniques indicate that SSN performs considerably better in terms of both 
ASA score and precision-recall. 
Fig.~\ref{fig:bsds_illustration}
shows a visual result comparing SSN$_{pix}$ and SSN$_{deep}$ and,
Fig.~\ref{fig:seg_visuals} shows visual results comparing 
SSN$_{deep}$ with state-of-the-arts. Notice that SSN$_{deep}$ superpixels
smoothly follow object boundaries and are also more concentrated near
the object boundaries.

\subsection{Superpixels for Semantic Segmentation}
\label{sec:semseg}

In this section, we present results on the semantic segmentation benchmarks of
Cityscapes~\cite{cordts2016cityscapes} and PascalVOC~\cite{everingham2015pascal}.
The experimental settings are quite similar to that of the previous section with
the only difference being the use of semantic labels as the pixel properties $R$
in the reconstruction loss. Thus, we encourage SSN to learn superpixels that adhere
to semantic segments.

%\noindent \textbf*{Cityscapes}
%\label{sec:cityscapes}

\setlength{\intextsep}{0pt}%no space at top
\begin{wraptable}[14]{r}{0pt}
  \tiny
  \begin{tabular}{>{\raggedright\arraybackslash}p{2.7cm}>{\centering\arraybackslash}p{1.2cm}>{\centering\arraybackslash}p{1.5cm}}
    \toprule
    \textbf{Model} & \textbf{GPU/CPU} & \textbf{Time (ms)}\\
    \midrule
    \tiny
  SLIC~\cite{achanta2012slic} & CPU & 350 \\
  SNIC~\cite{achanta2017snic} & CPU & 810 \\
  SEEDS~\cite{van2015seeds} & CPU & 160 \\
  LSC~\cite{li2015lsc} & CPU & 1240 \\
  ERS~\cite{liu2011entropy} & CPU & 4600 \\
  SEAL-ERS~\cite{Tu-CVPR-2018} & GPU-CPU & 4610\\
  GSLICR~\cite{ren2015gslicr} & GPU & 10 \\
  \midrule
  \emph{SSN models} &  & \\
  SSN$_{pix}$,v=10 & GPU & 58 \\
  SSN$_{deep}$,v=5,k=10 & GPU & 71\\
  SSN$_{deep}$,v=10,k=10 & GPU & 90\\
  SSN$_{deep}$,v=5,k=20 & GPU & 80\\
  SSN$_{deep}$,v=10,k=20 & GPU & 101\\
  \bottomrule
\end{tabular}
\mycaption{Runtime Analysis}{Average runtime (in ms) of different superpixel
techniques, for computing 1000 superpixels
on a $512 \times 1024$ cityscapes image.}

\label{tab:runtime}
\end{wraptable}

\begin{figure}[t]
\footnotesize
\hspace{0.75cm}%
\stackunder[0pt]{\includegraphics[width=5.0cm]{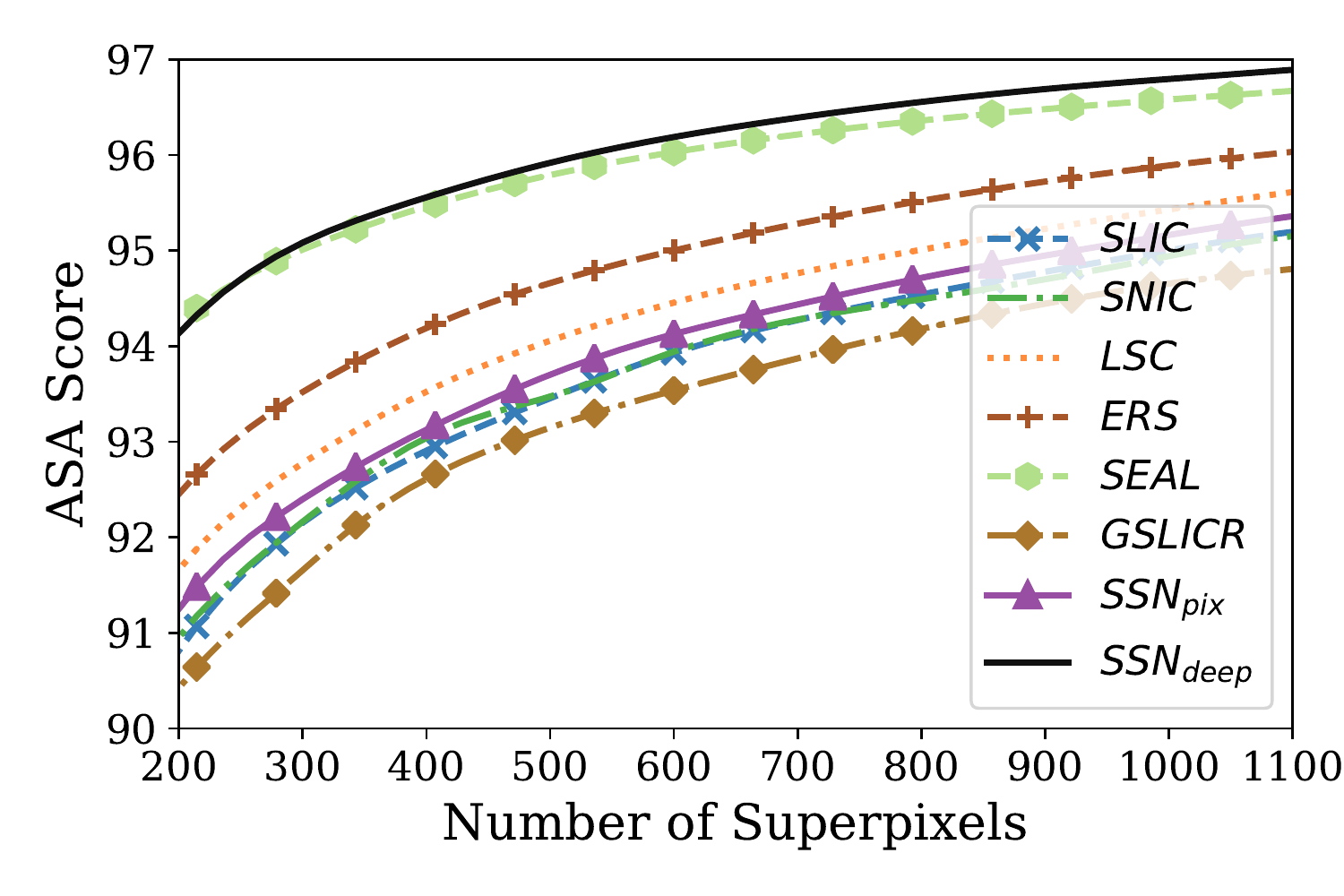}}{}%
\hspace{0.5cm}%
\stackunder[0pt]{\includegraphics[width=5.0cm]{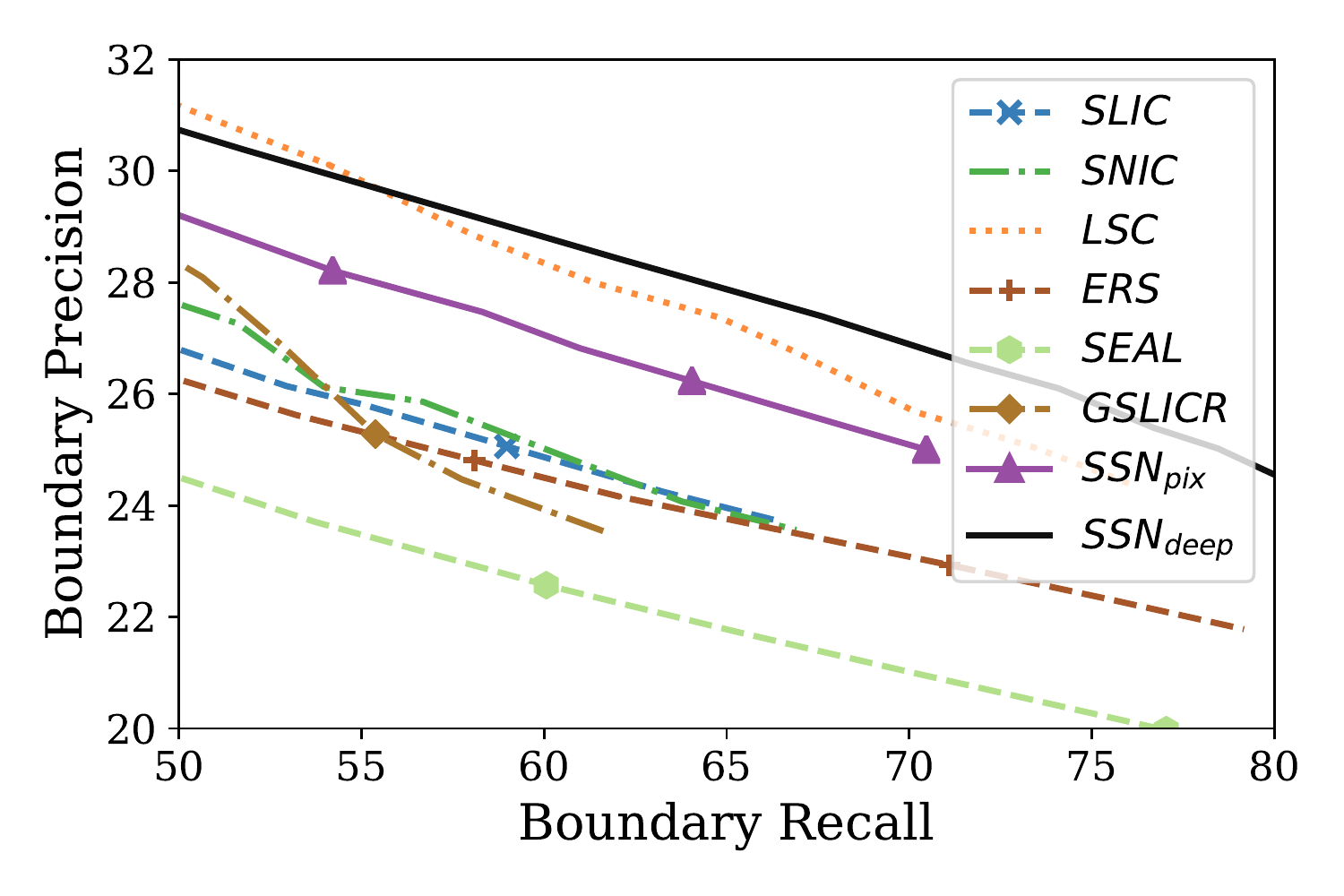}}{}
\mycaption{Results on Cityscapes validation}{ASA and boundary precision-recall shows that
SSN performs favourably against other techniques.}
\label{fig:scores_cs}
\end{figure}

\noindent \textbf{Cityscapes.}
Cityscapes is a large scale urban scene understanding benchmark with pixel accurate
semantic annotations. 
We train SSN with the 2975 train images and evaluate on the 500
validation images. 
For the ease of experimentation, we experiment with
half-resolution ($512 \times 1024$) images.
Plots in Fig.~\ref{fig:scores_cs} shows that SSN$_{deep}$ performs 
on par with SEAL~\cite{Tu-CVPR-2018} superpixels in terms of ASA
while being better in terms of precision-recall. 
We show a visual result
in Fig.~\ref{fig:seg_visuals} with more in the supplementary.

\noindent \textbf{Runtime analysis.} We report the approximate runtimes of different 
techniques, for computing 1000 superpixels on a $512 \times 1024$ 
cityscapes image in Table~\ref{tab:runtime}. We compute GPU runtimes using an
NVIDIA Tesla V100 GPU. The runtime comparison between SSN$_{pix}$ and SSN$_{deep}$
indicates that a significant portion of the SSN computation time is due to
the differentiable SLIC. 
%
% We believe that SSN can be made faster with a more
% optimized GPU implementation of differentiable SLIC.
%
The runtimes indicate that SSN is considerably faster than the implementations of 
several superpixel algorithms. 
% The only superpixel implementation
% that is faster than SSN is the GSLICR~\cite{ren2015gslicr},
% which is an approximate GPU implementation of SLIC.
% The plots in Fig.~\ref{fig:scores_cs} show that GSLICR performs much worse than SSN.

% \begin{wrapfigure}[15]{r}{0cm}
% \includegraphics[width=0.45\textwidth]{figures/voc12/asa_plot.pdf}
%   \mycaption{Results on PascalVOC} {ASA comparisons of SSN with other techniques
%   on the validation dataset.}
%   \label{fig:asa_voc12}
% \end{wrapfigure}

\begin{figure}[t]
\begin{center}
\centerline{\includegraphics[width=\columnwidth]{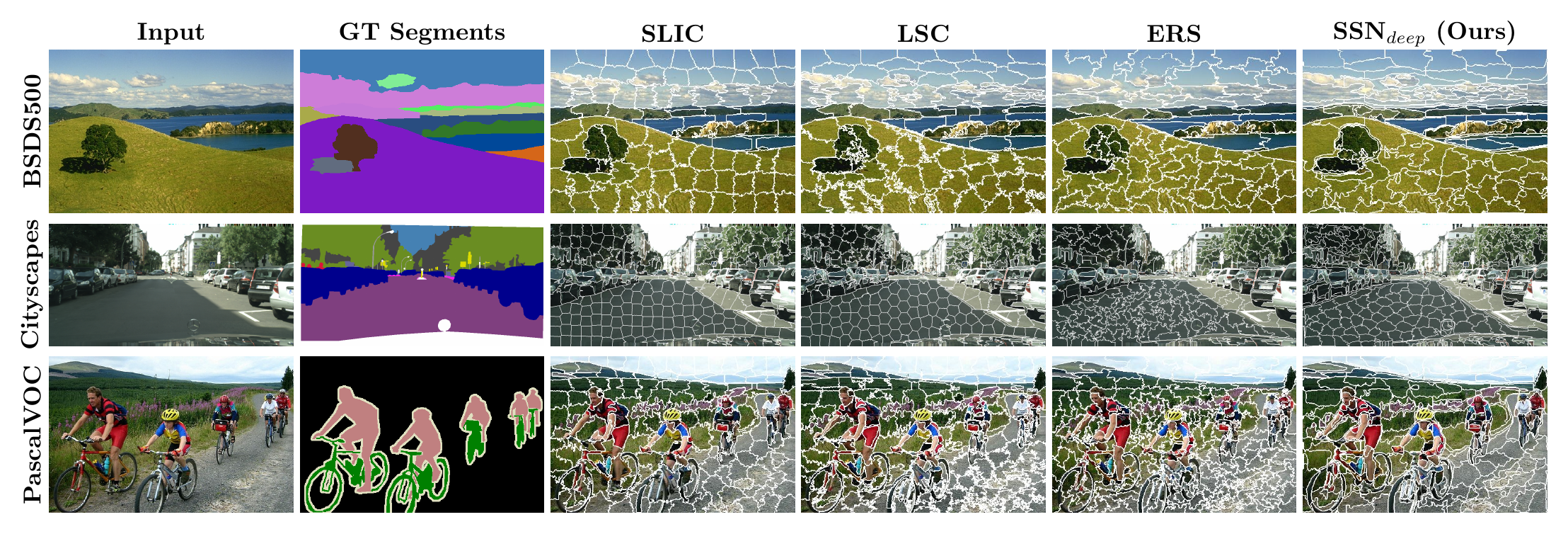}}
  \mycaption{Example visual results on different segmentation benchmarks}{Notice the segregation of
  SSN$_{deep}$ superpixels around object boundaries.}
  \label{fig:seg_visuals}
\end{center}
\end{figure}

\begin{figure}[t]
\footnotesize
\hspace{0.75cm}%
\stackunder[0pt]{\includegraphics[width=5.0cm]{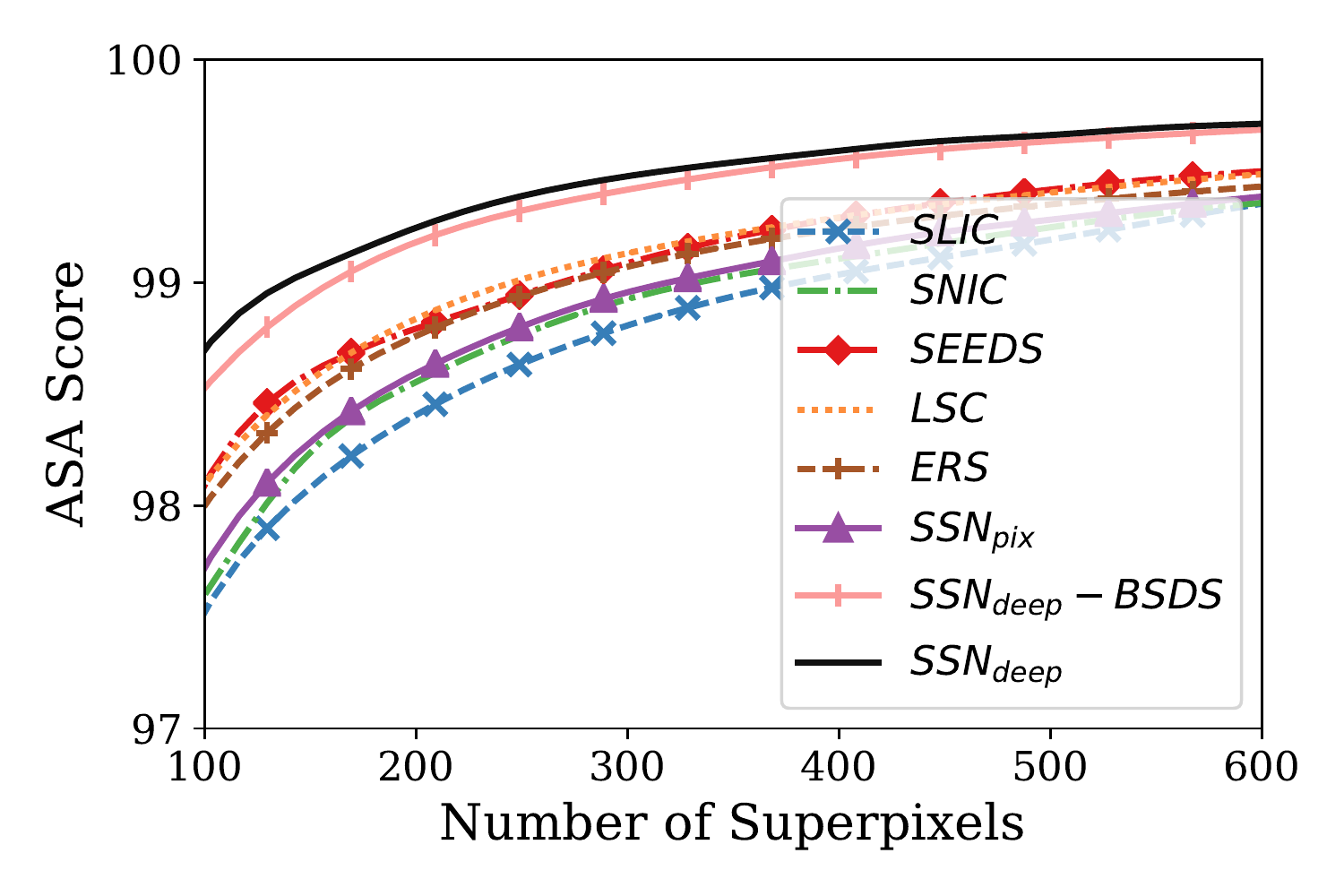}}
{\scriptsize (a) VOC Semantic Segmenation}%
\hspace{0.5cm}%
\stackunder[0pt]{\includegraphics[width=5.0cm]{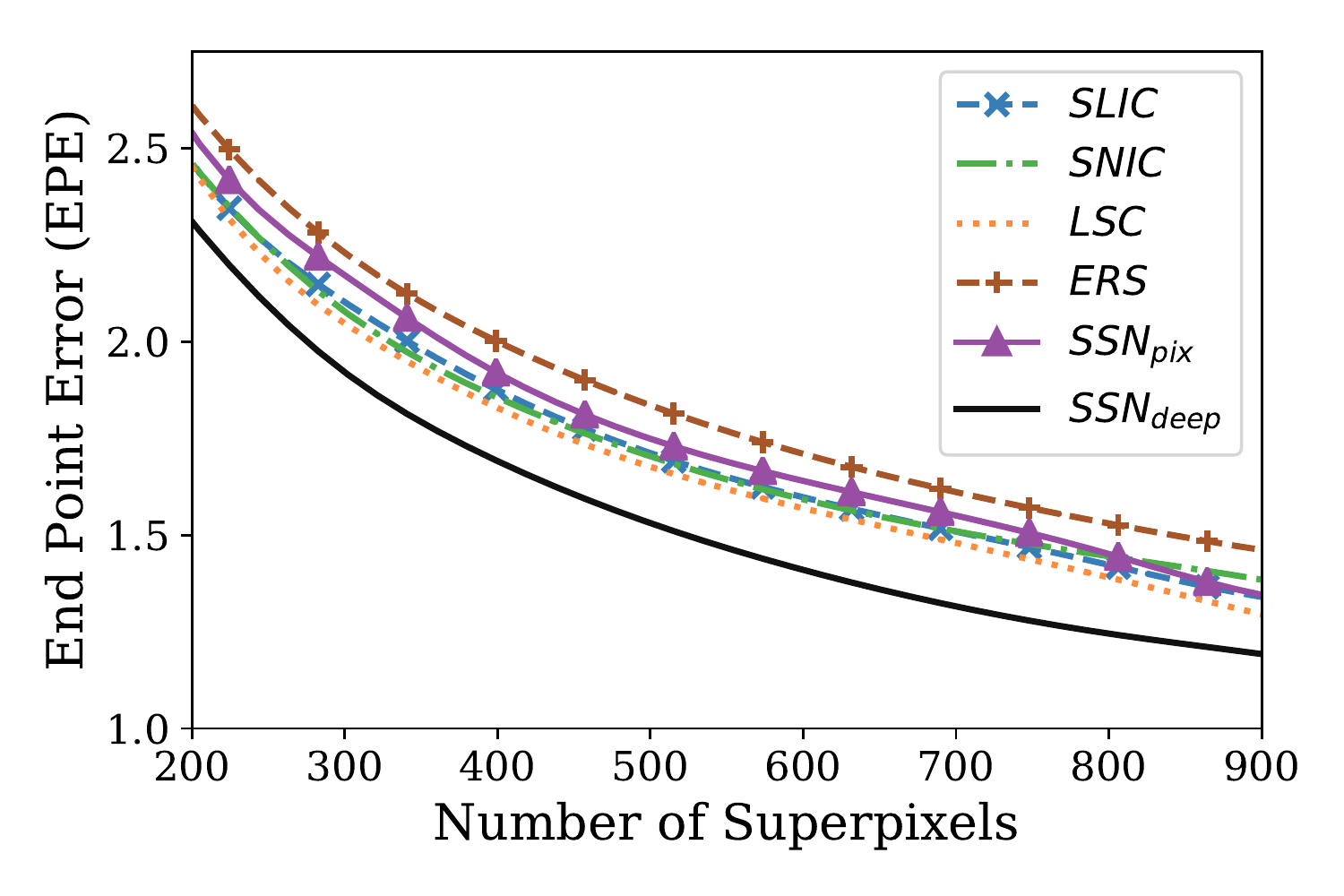}}
{\scriptsize (b) MPI-Sintel Optical Flow}%
\mycaption{Learning task-specific superpixels}{(a) ASA scores on PascalVOC2012
validation dataset and (b) EPE scores on Sintel optical flow validation dataset
showing the robustness of SSN across different tasks and datasets.}
\label{fig:scores_voc_sintel}
\end{figure}

%\noindent \textbf*{PascalVOC}
%\label{sec:voc12}

\noindent \textbf{PascalVOC.}
PascalVOC2012~\cite{everingham2015pascal} is another widely-used semantic 
segmentation benchmark, where we train SSN with 1464 train images and validate
on 1449 validation images. Fig.~\ref{fig:scores_voc_sintel}(a) shows the ASA scores
for different techniques. We do not analyze boundary scores on this dataset as the
GT semantic boundaries are dilated with an ignore label.
The ASA scores indicate that SSN$_{deep}$ outperforms other techniques.
We also evaluated the BSDS-trained model on this dataset and observed only a marginal
drop in accuracy (`SSN$_{deep}$-BSDS' in Fig.~\ref{fig:scores_voc_sintel}(a)).
This shows the generalization and robustness of SSN to different datasets. An example 
visual result is shown in Fig.~\ref{fig:seg_visuals} with more in the supplementary.

\setlength{\intextsep}{0pt}%no space at top
\begin{wraptable}[11]{r}{0pt}
  \scriptsize
    \begin{tabular}{p{3.0cm}>{\centering\arraybackslash}p{2.0cm}}
    % {@{}lcc@{}}
        \toprule
        Method &  IoU \\
        \midrule
        DeepLab~\cite{chen2015deeplab}              &   68.9 \\
        + CRF~\cite{chen2015deeplab}                 & 72.7 \\
        + BI (SLIC)~\cite{gadde16bilateralinception} & 74.1 \\
        + BI (SSN$_{deep}$)      &  \textbf{75.3} \\
        \bottomrule
    \end{tabular}
\mycaption{SSN with a downstream CNN}{IoU improvements, on the VOC2012 
test data, with the integration of SSN into the bilateral inception (BI) network from~\cite{gadde16bilateralinception}.}
\label{tab:ssn_bi}
\end{wraptable}

We perform an additional experiment where we plug SSN
into the downstream semantic segmentation network of~\cite{gadde16bilateralinception},
% which uses superpixels inside a CNN. 
The network in~\cite{gadde16bilateralinception}
has bilateral inception layers that makes use of superpixels for long-range
data-adaptive information propagation across intermediate CNN representations. 
Table~\ref{tab:ssn_bi}
shows the Intersection over Union (IoU) score
% , a standard evaluation metric for semantic segmentation, 
for this joint model evaluated on the test data. The
improvements in IoU with respect to original SLIC superpixels 
used in~\cite{gadde16bilateralinception} shows that SSN can also bring performance
improvements to the downstream task networks that use superpixels.

\subsection{Superpixels for Optical Flow}
\label{sec:flow}

To demonstrate the applicability of SSN for regression tasks as well, we
conduct a proof-of-concept experiment where we learn superpixels that adhere
to optical flow boundaries. 
% instead of object boundaries. 
To this end, we experiment
on the MPI-Sintel dataset~\cite{butler2012naturalistic} and 
use SSN to predict superpixels given a pair of input frames. 
We use GT optical flow as pixel properties $R$ in the reconstruction loss 
(Eq.~\ref{eqn:recon}) and use L1 loss for $\mathcal{L}$, encouraging SSN
to generate superpixels that can effectively represent flow.

\begin{figure}[t]
\begin{center}
\centerline{\includegraphics[width=\columnwidth]{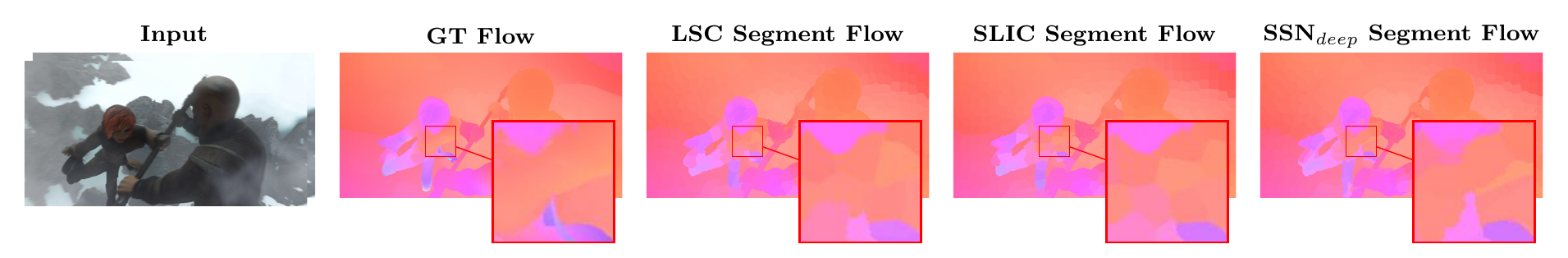}}
  \mycaption{Sample visual result on Sintel optical flow}{Segmented flow visuals obtained with different types of superpixels
  indicate that SSN$_{deep}$ superpixels can better represent GT optical flow compared to other techniques.}
  \label{fig:of_visuals}
\end{center}
\end{figure}

The MPI-Sintel dataset consists of 23 video sequences, which we split into
disjoint sets of 18 (836 frames) training 
and 5 (205 frames) validation sequences. To evaluate the superpixels,
we follow a similar strategy as for computing ASA. That is, for each
pixel inside a superpixel, we assign the average GT optical flow resulting
in a {\em segmented flow}. 
Fig.~\ref{fig:of_visuals} shows sample segmented flows obtained using different
types of superpixels. We then compute the Euclidean distance between
the  GT flow and the segmented flow, which is referred to as
end-point error (EPE). The lower the EPE value, the better the superpixels
are for representing flow. 
% Note that using a single flow value inside each
% superpixel is a restricted model for approximating optical flow.
% For practical applications, one would fit a more flexible affine model
% (similar to that in~\cite{yamaguchi2013robust}) to approximate flow using superpixels.
% In this work, for the sake of simplicity and
% for a direct comparison with other superpixel techniques, we use the simple
% averaging of flow inside superpixels.
A sample result in Fig.~\ref{fig:of_visuals} shows that SSN$_{deep}$
superpixels are better aligned with the changes in the GT flow than other superpixels.
Fig.~\ref{fig:scores_voc_sintel}(b) shows the average EPE values
for different techniques where SSN$_{deep}$ 
performs favourably against
existing superpixel techniques. This shows the usefulness of
SSN in learning task-specific superpixels.

\section{Conclusion}
\label{sec:conclusion}

%Summary of technique and contributions
We propose a novel superpixel sampling network (SSN) that 
% can generate high-quality superpixels
% from images. SSN 
leverages deep features learned via end-to-end training for
estimating task-specific superpixels. To our knowledge, this is the first deep superpixel
prediction technique that is end-to-end trainable. Experiments several benchmarks
show that SSN consistently performs favorably against 
state-of-the-art superpixel techniques, while also being faster.  
% This demonstrates the robustness of SSN
% across different datasets and tasks. 
Integration of SSN into a semantic
segmentation network~\cite{gadde16bilateralinception} also results in performance
improvements showing the usefulness of SSN in downstream computer vision
tasks. SSN is fast, easy to implement, can be easily integrated into other deep networks
and has good empirical performance.

% Future directions
SSN has addressed one of the main hurdles for incorporating superpixels into deep networks
which is the non-differentiable nature of existing superpixel algorithms.
The use of superpixels inside deep networks can have several advantages.
Superpixels can reduce the computational complexity, especially when processing high-resolution
images. Superpixels can also be used to enforce piece-wise constant assumptions and also help
in long-range information propagation~\cite{gadde16bilateralinception}. 
% in a single network layer.
%
We believe this work opens up new avenues in leveraging superpixels inside deep
networks and also inspires new deep learning techniques that use superpixels.

\small

\noindent \textbf{Acknowledgments.} We thank Wei-Chih Tu for providing evaluation scripts.
We thank Ben Eckart for his help in the supplementary video.

\clearpage

\bibliographystyle{splncs04}
\bibliography{refs}

\clearpage

\appendix

\section{Supplementary Material}

In Section~\ref{sec:metrics}, we formally define the
Acheivable Segmentation Accuracy (ASA) used for evaluating
superpixels. Then, in Section~\ref{sec:add_results},
we report F-measure and Compactness scores with more visual results on different datasets.
We also include a supplementary video\footnote{\url{https://www.youtube.com/watch?v=q37MxZolDck}}
that gives an overview of Superpixel Sampling Networks (SSN) with a glimpse of experimental results.

\subsection{Evaluation Metrics}
\label{sec:metrics}

Here, we formally define the Achievable Segmentation Accuracy (ASA) metric that is 
used in the main paper. Given an image $I$ with $n$ pixels, let $H \in 
\{0,1,\cdots,m\}^{n \times 1}$ denotes the superpixel segmentation with $m$ superpixels.
$H$ is composed of $m$ disjoint segments, $H = \bigcup_{j=1}^{m} H^{j}$, where
$j^{th}$ segment is represented as $H^j$. Similarly, let $G \in 
\{0,1,\cdots,w\}^{n \times 1}$ denotes ground-truth (GT) segmentation with $w$ segments.
$G = \bigcup_{l=1}^{w} G^{l}$, where $G^l$ denotes $l^{th}$ GT segment. 

\vspace{1mm}
\noindent \textbf{ASA Score.} The ASA score between a given superpixel segmentation $H$
and the GT segmentation $G$ is defined as

\begin{equation}
    ASA(H,G) = \frac{1}{n} \sum_{H^j \in S} \max_{G^l}|H^j \cap G^l|,
\end{equation}

where $|H^j \cap G^l|$ denotes the number of overlapping pixels between $S^j$ and
$G^l$. To compute ASA, we first find the GT segment
that overlaps the most with each of the superpixel segments and then sum the number of
overlapping pixels. As a normalization, we divide the number of overlapping pixels
with the number of image pixels $n$. In other words, ASA represents an upper bound
on the accuracy achievable by any segmentation step performed on the superpixels.

\vspace{1mm}
\noindent \textbf{Boundary Precision-Recall.} Boundary Recall (BR) measures how well the boundaries of superpixel segmentation aligns with the GT boundaries.
Higher BR score need not correspond to higher quality of superpixels.
Superpixels with high BR score can be irregular and may not be useful in practice.
Following reviewers' suggestions, we report Boundary Precision-Recall curves instead of 
just Boundary Recall scores. 

We also report F-measure and Compactness in the next section (Section~\ref{sec:compat_fscore}).  
We use the evaluation scripts from~\cite{stutz2016benchmark} 
with default parameters to compute Boundary Precision-Recall, F-measure and Compactness.

% BR is computed as:

% \begin{equation}
%     BR(H,G) = \frac{T(H,G)}{|B(G)|},
% \end{equation}

% where $|B(G)|$ represents the number of GT boundary pixels and $T(H,G)$ represents the
% number of boundary pixels in GT for which there is at least one superpixel boundary
% within range $r$. In our experiments, we set the range, $r=2$. 
% In simpler terms, BR represents the count of GT boundary pixels
% that fall within a close range of superpixel boundary pixels.

\subsection{Additional Experimental Results}
\label{sec:add_results}

\subsubsection{Compactness and F-measure.}
\label{sec:compat_fscore}

We compute compactness (CO) of 
different superpixels on the BSDS dataset (Fig.~\ref{fig:scores_bsds}(a)). SSN superpixels 
have only slightly 
lower CO compared to widely-used SLIC showing the practical utility of SSN. 
SSN$_{deep}$ has
similar CO as SSN$_{pix}$ showing that training SSN, while improving ASA and boundary 
adherence, does not destroy compactness. 
More importantly, we find SSN to be flexible and responsive to task-specific loss functions 
and one could use more weight ($\lambda$) for the compactness loss (Eq. 6 in the main paper) 
if more compact superpixels are desired.
In addition, we also plot F-measure scores in Fig.~\ref{fig:scores_bsds}(b). In summary,  SSN$_{deep}$ also outperforms other techniques in terms of F-measure while
maintaining the compactness as that of SSN$_{pix}$. This shows the robustness of SSN with
respect to different superpixel aspects.

\begin{figure}[t]
\footnotesize
\hspace{0.5cm}%
\stackunder[0pt]{\includegraphics[width=5.0cm]{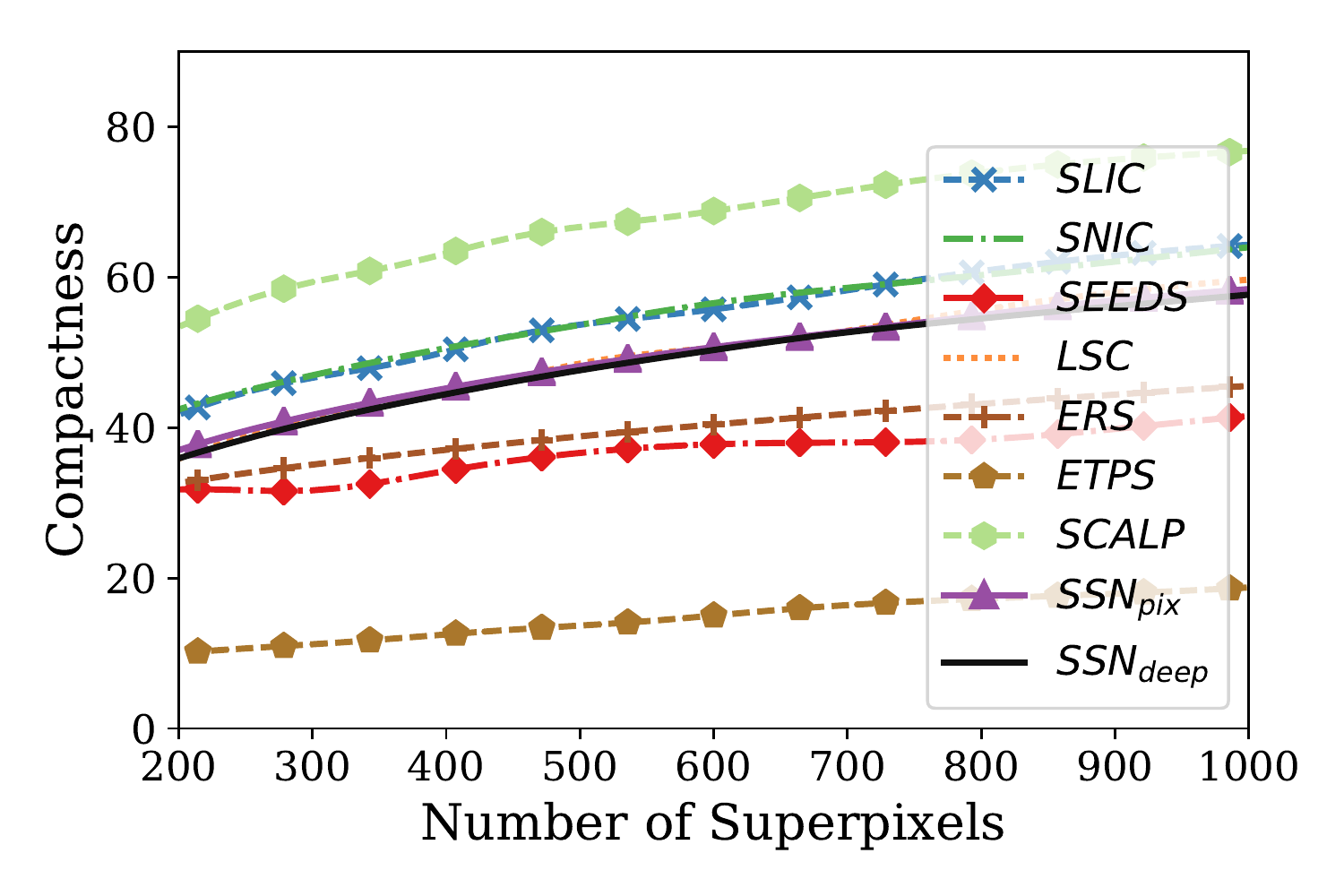}}
{\scriptsize (a) Compactness.}%
\hspace{1.0cm}%
\stackunder[0pt]{\includegraphics[width=5.0cm]{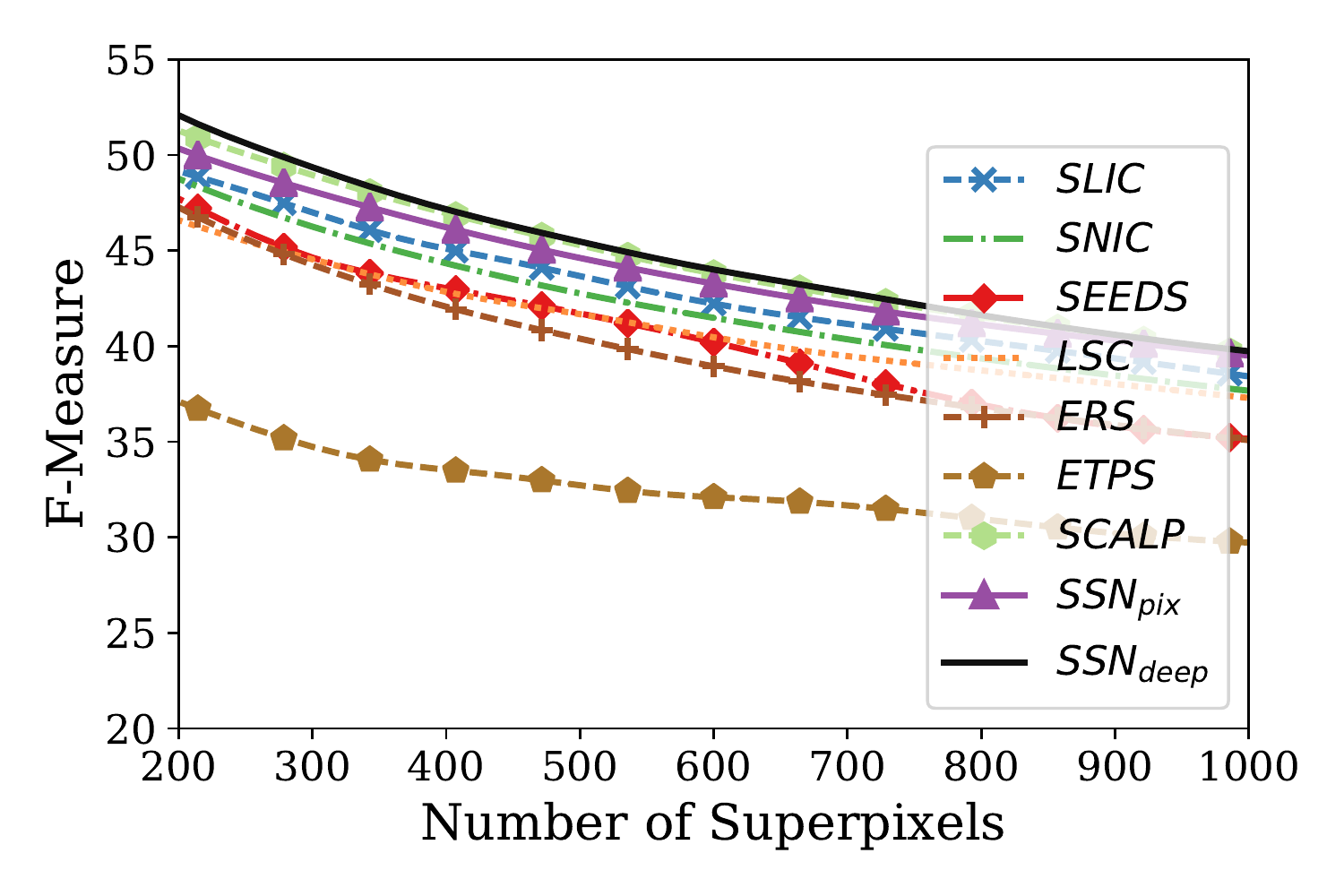}}
{\scriptsize (b) F-measure.}%
\mycaption{BSDS500 test results}{(a) Compactness and (b) F-measure plots. SSN$_{deep}$ outperforms
other techniques also in terms of F-measure while maintaing the same compactness as that of
SSN$_{pix}$.}
\vspace{5mm}
\label{fig:scores_bsds}
\end{figure}

\subsubsection{Additional visual results.}

\begin{figure}[t]
\begin{center}
\centerline{\includegraphics[width=\columnwidth]{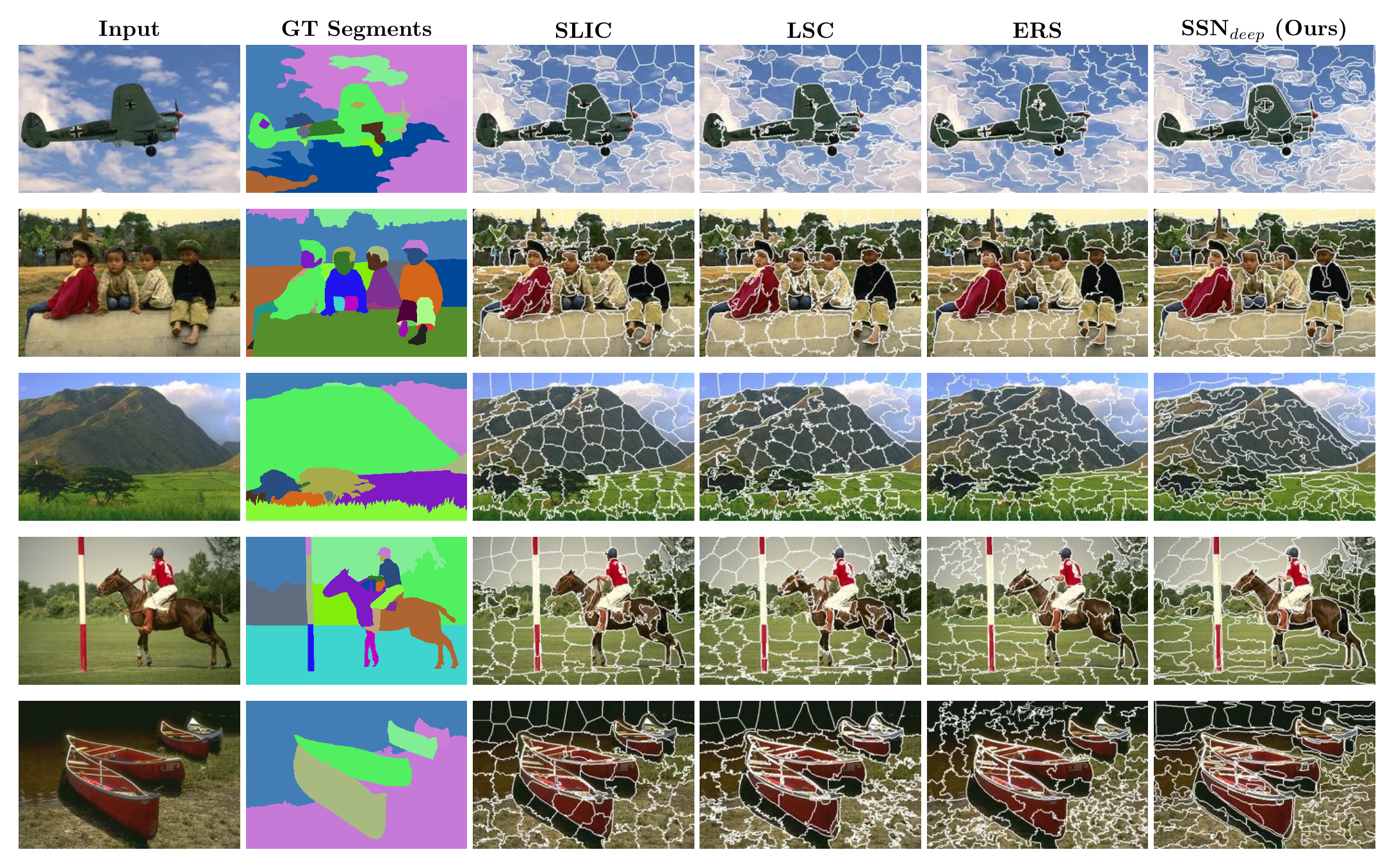}}
  \mycaption{Additional visual results on BSDS500 test images}{SSN$_{deep}$ tends to produce 
  smoother object contours and more superipxels near object boundaries in comparison to other
  superpixel techniques.}
  \label{fig:bsds_add_visuals}
\end{center}
\end{figure}

\begin{figure}[t]
\begin{center}
\centerline{\includegraphics[width=\columnwidth]{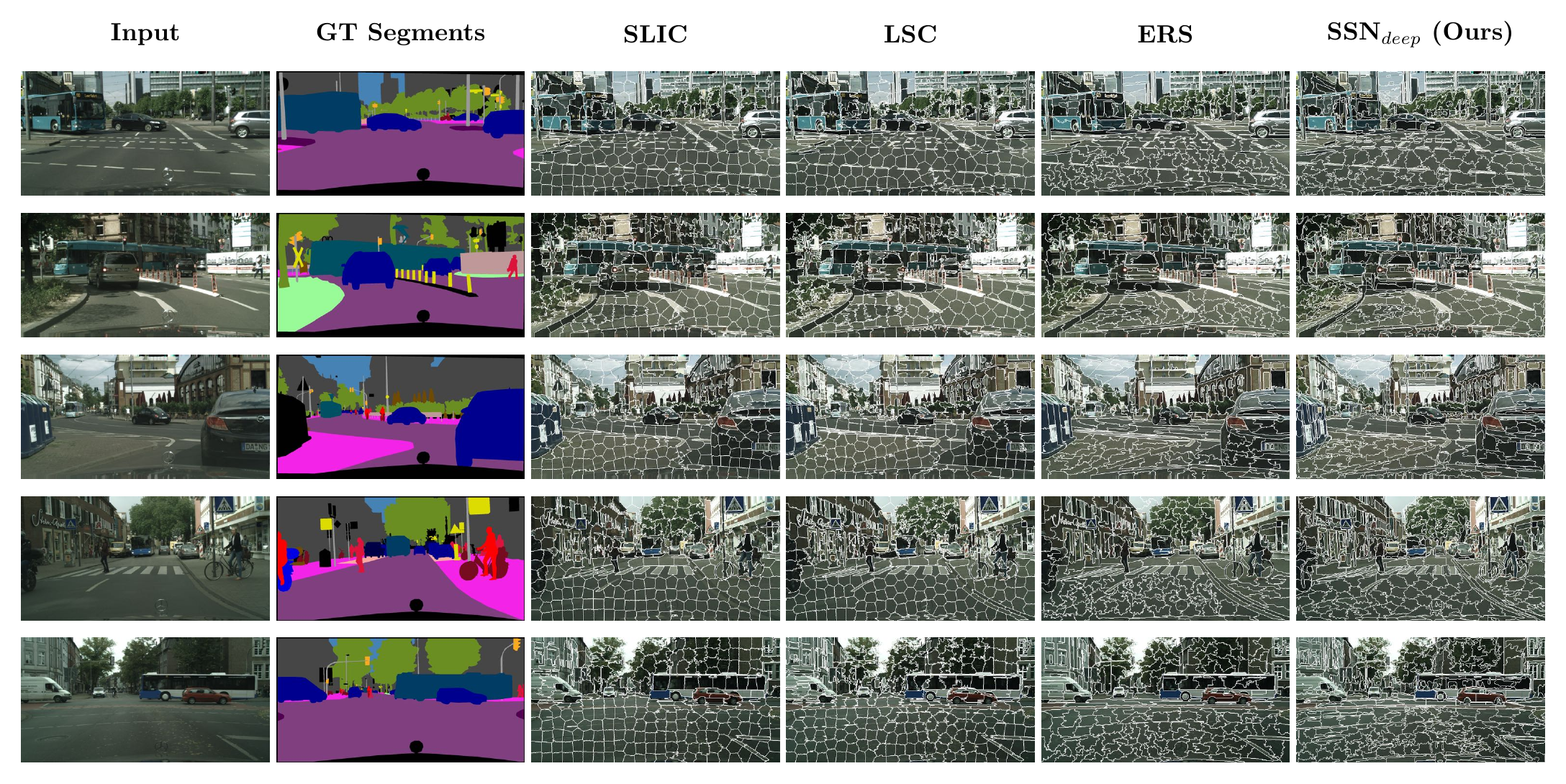}}
  \mycaption{Additional visual results on Cityscapes validation images}{SSN$_{deep}$ tend to
  generate bigger superpixels on uniform regions (such as road) and more superpixels on smaller
  objects.}
  \label{fig:cs_add_visuals}
\end{center}
\end{figure}

\begin{figure}[t]
\begin{center}
\centerline{\includegraphics[width=\columnwidth]{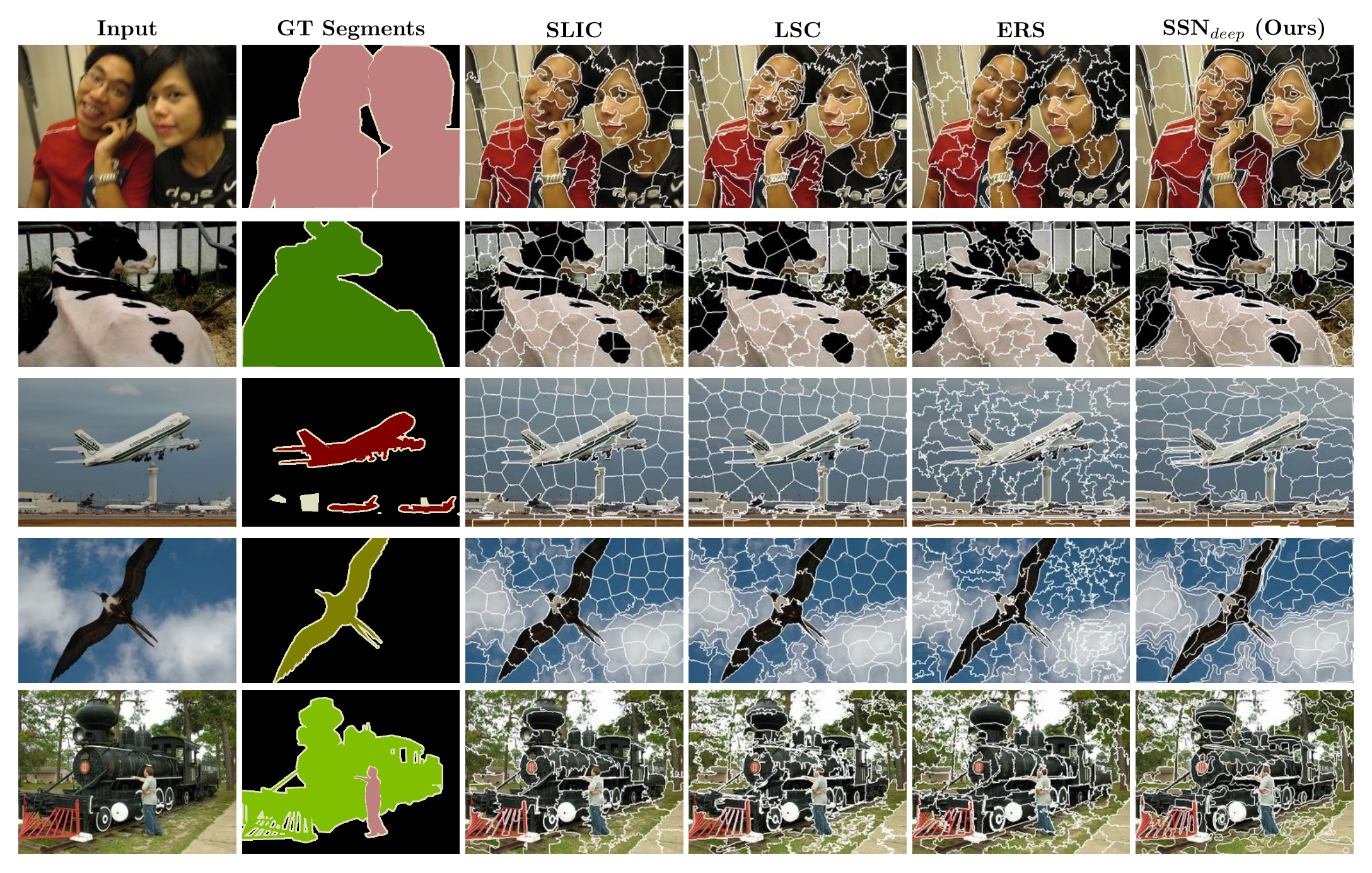}}
  \mycaption{Additional visual results on PascalVOC validation images}{SSN$_{deep}$ tends to
  produce smoother object contours and more superipxels near object boundaries in comparison to other
  superpixel techniques.}
  \label{fig:voc_add_visuals}
\end{center}
\end{figure}

\begin{figure}[t]
\begin{center}
\centerline{\includegraphics[width=\columnwidth]{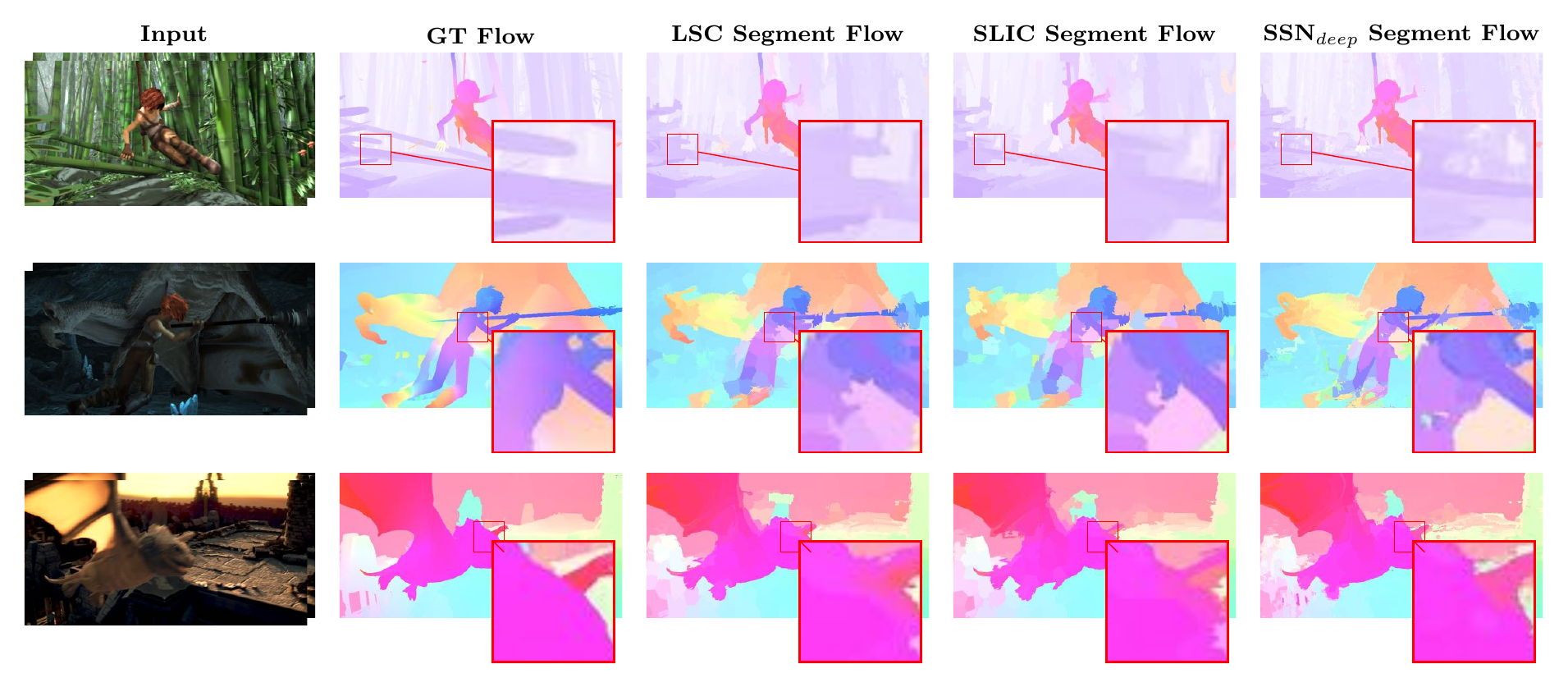}}
  \mycaption{Additional visual results on Sintel images}{Segmented flow visuals obtained with 
  different types of superpixels indicate that SSN$_{deep}$ superpixels can better represent GT optical 
  flow compared to other techniques.}
  \label{fig:of_add_visuals}
\end{center}
\end{figure}

In this section, we present additional visual results of different techniques and on different
datasets. Figs.~\ref{fig:bsds_add_visuals},~\ref{fig:cs_add_visuals} and~\ref{fig:voc_add_visuals} show superpixel visual results
on three segmentation benchmarks of BSDS500~\cite{amfm2011bsds}, 
Cityscapes~\cite{cordts2016cityscapes} and PascalVOC~\cite{everingham2015pascal} respectively.
For comparisons, we show the superpixels obtained with 3 existing superpixel techniques of
SLIC~\cite{achanta2012slic}, LSC~\cite{li2015lsc} and ERS~\cite{liu2011entropy}.
Fig.~\ref{fig:of_add_visuals} shows additional visual results on MPI-Sintel~\cite{butler2012naturalistic}
where we present sample segmented flows obtained using different types of superpixels.

\end{document}